\documentclass[journal]{IEEEtran}

\usepackage{soul}
\usepackage{hyperref}
\hypersetup{
    colorlinks=true,
    linkcolor=blue,
    filecolor=magenta,      
    urlcolor=cyan,
    pdftitle={Overleaf Example},
    pdfpagemode=FullScreen,
    }
    
\ifCLASSINFOpdf
   \usepackage[pdftex]{graphicx}
   \DeclareGraphicsExtensions{.pdf,.jpeg,.png}
\else
  \usepackage[dvips]{graphicx}
\fi

\usepackage{amsmath}

\usepackage{multirow}
\usepackage[table,xcdraw]{xcolor}

\usepackage[table,xcdraw]{xcolor}

\ifCLASSOPTIONcompsoc
  \usepackage[caption=false,font=normalsize,labelfont=sf,textfont=sf]{subfig}
\else
  \usepackage[caption=false,font=footnotesize]{subfig}
\fi

\newcolumntype{L}[1]{>{\raggedright\let\newline\\\arraybackslash\hspace{0pt}}m{#1}}
\newcolumntype{C}[1]{>{\centering\let\newline\\\arraybackslash\hspace{0pt}}m{#1}}
\newcolumntype{R}[1]{>{\raggedleft\let\newline\\\arraybackslash\hspace{0pt}}m{#1}}

\usepackage{url}

\begin{document}
%
\title{Diegetic Graphical User Interfaces \& \\ Intuitive Control of Assistive Robots via Eye-gaze}

%

\author{Emanuel~Nunez Sardinha${}^{1}$,~\IEEEmembership{Member,~IEEE,}
Marcela~Múnera${}^{1}$,~\IEEEmembership{Fellow,~OSA,}
Nancy~Zook${}^{2}$,~\IEEEmembership{Fellow,~OSA,}
David~Western${}^{1}$,~\IEEEmembership{Fellow,~OSA,}
and~Virginia~Ruiz Garate${}^{3}$,~\IEEEmembership{Member,~IEEE}
        
\thanks{$1$ Is with the Bristol Robotics Laboratory and University of the West of England (UWE), Bristol, United Kingdom}
\thanks{$2$ N. Zook is with Psychological Sciences Research Group, School of Social Sciences, University of the West of England (UWE), Bristol, United Kingdom.}
\thanks{$3$ V. Ruiz Garate is with Mondragon Unibertsitatea - Faculty of Engineering and the University of the West of England vruiz@mondragon.edu}
\thanks{$*$ Corresponding author Emanuel.NunezSardinha@uwe.ac.uk} 
}

%
%

\markboth{Journal of \LaTeX\ Class Files,~Vol.~14, No.~8, August~2015}%
{Shell \MakeLowercase{\textit{et al.}}: Bare Demo of IEEEtran.cls for IEEE Journals}
%



\maketitle

\begin{abstract}


Individuals with tetraplegia and similar forms of paralysis suffer physically and emotionally due to a lack of autonomy. 
To help regain part of this autonomy, assistive robotic arms have been shown to increase living independence. 
However, users with paralysis pose unique challenging conditions for the control of these devices.  
In this article, we present the use of \textit{Diegetic Graphical User Interfaces}, a novel, intuitive, and computationally inexpensive approach for gaze-controlled interfaces applied to robots.
By using symbols paired with fiducial markers, interactive buttons can be defined in the real world which the user can trigger via gaze, and which can be embedded easily into the environment.   
We apply this system to pilot a 3-degree-of-freedom robotic arm for precision pick-and-place tasks. 
The interface is placed directly on the robot to allow intuitive and direct interaction, eliminating the need for context-switching between external screens, menus, and the robot.
After calibration and a brief habituation period, twenty-one participants from multiple backgrounds, ages and eye-sight conditions completed the Yale-CMU-Berkeley (YCB) Block Pick and Place Protocol to benchmark the system, achieving a mean score of $13.71$ out of the maximum $16.00$ points. 
Good usability and user experience were reported (System Usability Score of $75.36$) while achieving a low task workload measure (NASA-TLX of $44.76$). 
Results show that users can employ multiple interface elements to perform actions with minimal practice and with a small cognitive load.
To our knowledge, this is the first easily reconfigurable screenless system that enables robot control entirely via gaze for Cartesian robot control without the need for eye or face gestures. 

\end{abstract}

%
\IEEEpeerreviewmaketitle

\section{Introduction}


\IEEEPARstart{T}{etraplegia} is the total or partial paralysis of all limbs
affecting patients' quality of life, impairing mobility, sensing, and communication \cite{InternationalStandardsNeurologicalClassificationsSpinalCordInjury_2011, TetraplegiaManagementUpdate_Friden}.
The condition requires extensive and life-long physical and psychological support, depending on external assistance to perform most of the \textit{Activities of Daily Living} (ADLs) \cite{AComparisonofTwoFunctionalTestsinQuadriplegiaTheQuadriplegiaIndexofFunctionandtheFunctionalIndependenceMeasure_Yavuz}. 
Over the past 40 years, the prognosis of prenatal and traumatic conditions resulting in tetraplegia has improved and life expectancy has been increasing, now sitting at 38 years post-injury in the northern hemisphere \cite{IncidencePrevalenceEpidemiologySpinalCordInjuryWorldwideLiteratureSurvey_Wyndaele,AComparativeReviewofLifeSatisfactionQualityofLifeandMoodBetweenChineseandBritishPeoplewithTetraplegia_Songhuai}.
However, this has also increased the number of patients requiring long-term care, further straining the healthcare system \cite{ComparisonTwoFunctionalTestsQuadriplegia_Yavuz}. 
Individuals with tetraplegia are often forced to depend on family and friends \cite{TheHealthandLifePrioritiesofIndividualswithSpinalCordInjuryReview_Simpson}, which in turn also places them under considerable financial and emotional burden.
This not only impacts the quality of life of the people with tetraplegia and their close-ones others but also translates into a significant financial burden, with the average lifetime cost in the UK for injury-related tetraplegia of \pounds 1.12 million predominantly in care costs \cite{UnderstandingandModellingtheEconomicImpactofSpinalCordInjuriesintheUnitedKingdom_McDaid}, for a total of \pounds 1.43 billion per year.
Moreover, people with tetraplegia present diverse aetiology, symptoms, and evolution, placing personalised requisites on their control systems \cite{orejuela-zapataSelfHelpDevicesQuadriplegic2019} and making one-size-fits-all solutions limited in reach.
Individuals with tetraplegia require robust but personalisable systems that can continuously adapt to their physical symptoms and requirements, cognitive capabilities, environment, and preferences.

Recently, efforts in tetraplegia care have refocused on improving \textit{patient autonomy}, associated with greater emotional health, employment, and life satisfaction \cite{AgeingWithSpinalCordInjury-CrosssectionalandLongitudinalEffects_Weitzenkamp, HealthLifePrioritiesIndividualsSpinalCordInjury_Review_Simpson}. 
Restoration of upper limb function, particularly of the hand, appears in most studies as one of the three most important health priorities for patients with tetraplegia  \cite{TheHealthandLifePrioritiesofIndividualswithSpinalCordInjuryReview_Simpson}.  
Thus, hand function has been attempted to be reinstated via a variety of assistive tools \cite{AccesstoEquipmentParticipationandQualityofLifeinAgingIndividualswithHighTetraplegia_Bushnik} and surgical procedures \cite{TetraplegiaManagementUpdate_Friden}. 
For people with partial paralysis and advanced forms of reduced mobility, the use of a joystick-controlled wheelchair and robotic arm has been reported as one the most effective tools to surmount environmental barriers and develop independence, reduce care time, and improve quality of life \cite{LongtermUseJACO_Beaudoin,EvaluationJACOClinicoEconomicStudyWheelchairUpperExtremityDisabilities_Maheu} 
However, adapting these systems to people with tetraplegia while maintaining a good and personalised user experience, has proven to be an exceedingly challenging design task, especially when robots can exceed the 6 Degrees of Freedom (DOF). 
This is further complicated as assistive devices for tetraplegia are usually abandoned, perceived as too expensive, slow, unhelpful, exhausting to use, unattractive to users \cite{orejuela-zapataSelfHelpDevicesQuadriplegic2019}, or depending on hardware, causing feelings of alienation or abnormality \cite{AssessmentofBrainMachineInterfacesfromthePerspectiveofPeoplewithParalysis_Blabe}.

Multiple types of physiological instrumentation have been used to drive assistive devices, 
including EEG \cite{hochbergReachGraspPeople2012a}, 
motion tracking \cite{leeExploratoryMultiSessionStudy2023,rudigkeitAMiCUSHeadMotionbased2019}, 
tongue control \cite{palsdottirDedicatedToolFrame2022a}, 
speech recognition, 
and facial gesture recognition \cite{noronhaWinkGraspComparing2017}.
However, none have reached widespread adoption due to complexity, invasiveness, or poor user experience. 
Eye-gaze control stands out as a powerful alternative interface for the population with tetraplegia, as individuals usually retain good control over their eyes, even in long-term degenerative conditions \cite{kaminskiDifferentialSusceptibilityOcular2002}. 
Compared to the popular Brain-Computer Interfaces (BCI), eye-gaze detection through eye-tracking needs less training time, is faster to calibrate, less invasive, less complex, and less expensive \cite{GazebasedTeleprosthetic_Dziemian}. 
The use of gaze-controlled on-screen interfaces has also been adopted by successful commercial solutions to tackle communication \cite{Grid3,TobiiDynavox} and wheelchair driving \cite{Eyedrivomatic,MyEccControl}.
Yet, similar approaches for robot control have not reached a suitable readiness level, with systems fully controlled by the user being complex, slow, unintuitive, and fatiguing, or autonomous options being limited to structured scenarios (see Section \ref{s:RelatedWorks}). 
Thus, there is a distinct need for robust gaze-control interfaces for wheelchair-mounted robotic arms that are capable of executing precise and swift movement, while at the same time remaining intuitive and simple \cite{huangEOGbasedWheelchairRobotic2019}.

This manuscript presents a new intuitive and re-configurable interface system based on eye-gaze interaction informed by the advances in literature and the needs and limitations of people with tetraplegia. 
It expands gaze control outside of screens, mitigates shortcomings associated with on-screen or augmented reality devices such as context switching and nausea \cite{gabbardEffectsARDisplay2019}, and sets the stage for easy user customisation. 
The \textit{Diegetic Graphical User Interface} (D-GUI) demonstrates a novel and computationally inexpensive approach for a screenless GUI in which simple symbols can be arranged into a GUI embedded in the robot. 
We demonstrate the developed system for continuous cartesian control of a robot arm for precision tasks, benchmarked with the standardised \textit{Yale-CMU-Berkeley (YCB) Block Pick and Place Protocol} \cite{BenchmarkingManipulationResearchUsingYCBObjectandModelSet_Calli_2015}, with users finding the system intuitive and producing low-task workloads. 

In Section \ref{s:RelatedWorks} we briefly discuss previous eye-gaze interface approaches for controlling assistive robotic arms, with emphasis on systems for individuals with tetraplegia, their back-end technology and readiness level. 
In Section \ref{s:DiegeticUserInterfaces} the working principle and implementation of the proposed system is explained, together with some basic performance measurements. 
Section \ref{s:RobotControl} discusses its application for robot control, our interface and control method. 
Section \ref{s:Methods} discusses the benchmarking methodology for the system, based on the standardised YCB pick and place test, while Section \ref{s:Results} presents the results for the benchmarking methods and Section \ref{s:Discussion} our discussion over the results. 
In Section \ref{s:Conclusion} we present our conclusions and plans for future works. 
Finally in Annex \ref{s:OpenSourcePackages} we present the used software packages, available as open-source software for easy reproduction and experimentation.


\section{Related Works}
\label{s:RelatedWorks}

Tetraplegia can have diverse aetiology, caused by 
traumatic accidents \cite{TetraplegiaManagementUpdate_Friden} (Spinal Cord Injury, Traumatic Brain Injury), 
body or neuro-degenerative diseases (Cerebral Palsy, Amyotrophic Lateral Sclerosis, Multiple Sclerosis, Muscular Dystrophy Duchenne) or
strokes. 
Regardless of root condition, eye-gaze-based control is a powerful alternative interface for assistive tools, as individuals usually retain good control over their eyes \cite{kaminskiDifferentialSusceptibilityOcular2002}, even when the ability to speak is affected.
Compared to EEG systems, control via eye-tracking needs less training time, is faster to calibrate, less invasive, and cheaper to implement \cite{GazebasedTeleprosthetic_Dziemian}. 
Gaze control is achieved via gaze detection, where \textit{gaze} refers to the point in space a person is looking at, defined as a 3D \textit{Point of Interest} (POI) or \textit{gaze point} estimated from the position and rotation of the user's eyes \cite{li3DGazeBasedRoboticGrasping2017}. 
These are different from eye gestures, a series of eye or facial movements used to trigger actions. 
While useful, using these in excess can lead to exhaustion \cite{orejuela-zapataSelfHelpDevicesQuadriplegic2019}. 
This section covers the state of related literature regarding the control of robotic systems using primarily gaze organized by approach. 
Emphasis is placed on control methods and user experience, presented in the context of assistance to people with tetraplegia.


\subsection{Graphical User Interfaces}


A common method of gaze control of assistive arms is the use of Graphical User Interfaces (GUIs) to activate the different functions of a robot. 
Interaction is performed with a screen, where the gaze point is projected onto two-dimensional screen coordinates \cite{li3DGazeBasedRoboticGrasping2017}. 
Gaze then acts analogous to a mouse cursor and is used to select from a pool of actions for the robot to execute.
Kim et al. \cite{kimHumanRobotInterfaceUsing2001} demonstrate the earliest of these approaches, achieving direct joint control of a 4DOF robot via a screen interface where the user fixates on a UI button corresponding to the desired joint and then uses pop-up menus to move the robot in discrete joint space. 
A camera view of the robot is provided to the user simultaneously.  
To prevent accidental triggering of the commands, a brief dwell time window is also applied. 
A similar strategy is adopted by Di Maio et al. \cite{dimaioHybridManualGazeBased2021}, using sliders instead of buttons for movement, while also enabling scrolling through menus via moving the eyes left and right.
The use of joint space controls is criticized by Alsharif et al. \cite{alsharifGazeGestureBasedHuman2016}, as it offloads the inverse kinematic calculations to the user leading to a higher cognitive load. 
In the \textit{MyECC Pupil} by Homebrace \cite{MyEccPupil}, to our knowledge the only commercial device capable of allowing control of robotic systems for people with tetraplegia, the user can control the robot end-effector position and rotation in cartesian space by interacting with a GUI.
The user can also controll the end effector by looking in directions relative to the glasses, i.e. look left to move the end effector left.
However, a maximum of 2 DOF can be manipulated at once, requiring switching modes to use other functions, such as depth control, rotation and the different grasping modes. 
While the device enables control of all available DOF, depending on the end-effector position it might require the user to look away from the end effector being controlled, introducing safety and usability concerns. 
Pure screen interaction also removes the user's attention from the world \cite{wangFreeView3DGazeGuided2018}, increasing the likelihood of errors. 
Furthermore, as explored in augmented reality applications, the frequent context-switching between visual and cognitive attention from the screen to the real-world robot requires adjusting attention and focal distance, leading to increased eye strain, affecting usability, increasing fatigue and contributing to simulator sickness \cite{gabbardEffectsARDisplay2019}.

Camera feeds can be included on the GUI, providing the user with views \textit{of} the robot or \textit{from} the robot.  
Fuji et al. \cite{fujiiGazeContingentCartesian2013} presents a context-sensitive UI for control of a laparoscopic camera, with the control interface overlaid on top of a camera feed from the camera on the robot. 
The user thus controls from the point of view of the end-effector, switching between different camera modes (zoom and pan) using gestures, or changing the control mode by looking at different areas of the UI. 
In these examples the small screen space available limits the number of interactable items, requiring navigation through menus to achieve control, or the number of options has been intentionally reduced to simplify its user. 
Park et al. \cite{parkHandsFreeHumanRobot2021} adapts the GUI strategy to Mixed Reality systems where a floating GUI is displayed next to the robot when interacting with an item. 
Using gaze, the user can select the end effector and control position or rotation over cartesian space in discrete amounts. 
The approach allows menu-based interaction without the size constraints of a screen and can preview the robot status and target positions in real-time. 
However, gaze interaction is still limited to coarse control, with head movements used as the primary control method for precise tasks, and menus are still needed to switch between the different interaction modes. 
While the use of an Augmented Reality headset provides useful feedback, its size and need for frequent head movements make it less suitable as an everyday tool for a population with tetraplegia. 
As per Blabe et al. \cite{AssessmentofBrainMachineInterfacesfromthePerspectiveofPeoplewithParalysis_Blabe}, bulky devices may also introduce an aesthetic barrier, making them less desirable, but does suggest their use in more discrete form factors. 
The use of GUIs for interaction has been criticised as leading to slow behaviour and extensive screen coverage in gaze-controlled surgical applications \cite{fujiiGazeContingentCartesian2013,fujiiGazeGestureBased2018}. 
To make interactions clear, buttons on menus have to be made large relative to screen space, reducing the number of available options. 
When combined with the wait introduced by dwell times, menus significantly slow down interaction, which can lead to poor user experience.



Beyond of direct servoing of the robot other input strategies have been demonstrated via GUIs. 
Fortini et al. \cite{fortiniCollaborativeRoboticApproach2019} demonstrates eye control for teaching actions to an industrial robot arm, tracing the path set by the user's hand, and triggering multiple automated actions via gaze.
Using an eye-tracking bar and hand-drawn buttons, the user can trigger the different points in the trajectory, execute it, or move the arm to a rest position, but this does not address the actual control strategy. 
In the ANSOS multi-stage experiments \cite{laffontEvaluationGraphicInterface2009a,ANSOStudyEvaluationinanIndoorEnvironmentofaMobileAssistanceRoboticGraspingArm_Coignard} individuals with different forms of tetraplegia tested a robot to automatically fetch items using a graphical interface. 
The user traces a square shape with their gaze to mark a bounding box for the item, setting the target path for the robot to pick up. 
If the robot was unable to autonomously find the item, the user would take over remotely and use buttons to directly move the arm along 2 DOF relative to the viewport. 
Despite the technical success, only 25\% of users reported being interested in using the robot for everyday tasks \cite{ANSOStudyEvaluationinanIndoorEnvironmentofaMobileAssistanceRoboticGraspingArm_Coignard}. 
Overall, GUI systems provide an opportunity for control of complex robotic systems, providing means for immediate feedback and task confirmation. 
However, the reduced screen space in portable screens means most mobile systems feature very simple interfaces, resulting in a slow user experience \cite{fujiiGazeGestureBased2018} and requiring navigating to multiple menus.



\subsection{End-point control}


The \textit{visual axes intersection} approach is identified as one of the most popular and accurate and involves calculating an axis passing normally through each pupil and defining the intersection point as a 3D \textit{Point of Interest}.

Other approaches focus instead on the direct interaction between the user's gaze, the robot, and the space around them, omitting the GUI. 
A common approach is to instruct the robot to reach the 3D point of gaze to move the end effector to where the user is looking. 
Li et al. \cite{li3DGazeBasedRoboticGrasping2017} cover the evolution of methods for \textit{three-dimensional} gaze point reconstruction, of which the visual axis intersection method is the most popular. 
This method estimates the intersection between both eyes' vergence, approximating the intersection as a 3D gaze point using the least square error method. 
However, this leads to inaccurate depth, as any errors in estimating the axes propagate and amplify, and thus robot control strategies must take measures to minimize the error. 
In \cite{3DGazeCursor_Tostado}, 3D continuous movement of the end-effector to grasp items is demonstrated, with the user winking to trigger opening or closing. 
However, the approach requires resting the user's face fixed in place very close proximity to the end effector to increase accuracy, resulting in limited workspace and usability. 
The follow-up work \cite{maimon-morFree3DEndpoint2017} reworks the design as an exoskeleton control system, moving the participant's arm by looking at the end target and using winks to confirm selection. 
It improves usability by expanding the interaction space to a wider 60cm x 70cm area, and thanks to real-time head tracking the user can freely move their head during the operation. 
However, the expanded area reduces precision, and thus end-effector movement is discretised into set regions of space.
The interaction is further restricted to a 2D space above the interaction, and significantly covers the face of the user, introducing a known barrier for adoption \cite{AssessmentofBrainMachineInterfacesfromthePerspectiveofPeoplewithParalysis_Blabe}. 
A similar approach is followed by Dziemian et al. \cite{GazebasedTeleprosthetic_Dziemian} showing intuitive continuous 3D control of a robot for drawing, where gaze controls 2D movement on a flat surface and a combination of winks and dwell time is used to filter and confirm choices. 
Head tilts are necessary to control depth, however, and the interaction has to occur orthogonal to the user for best results. 
While pure end-point control is foundational, the 3D control of the end effector via gaze alone presents unsolved limitations, like significantly reduced or structured workspaces \cite{3DGazeCursor_Tostado, maimon-morFree3DEndpoint2017}, lack of feedback, discretized movement \cite{maimon-morFree3DEndpoint2017}, and requiring (rather than benefiting from) gestures or multimodal commands \cite{maimon-morFree3DEndpoint2017,GazebasedTeleprosthetic_Dziemian}.  
It must be supported by automated or complementary systems to maximise usability, allowing the robot to decide how to tackle the grasping of ambiguous, translucent, occluded, or out-of-view objects.

\subsection{Real-world interaction}



More recent attempts focus on automated measures to simplify control for the user and minimize error, expanding end-point control. 
Direct interaction with the items within a scene is common, where the user selects objects naturally via gaze, either through a screen or in the real world. 
The robot will then autonomously decide how to navigate and interact with the objects, using manipulation techniques from traditional robotics. 
Different back-ends and gaze-filtering technologies have been demonstrated following this strategy, using gaze fixation \cite{wangFreeView3DGazeGuided2018, cioProofConceptAssistive2019, fedorovaGazeBasedRobot2015}, hitscan, and Earth Mover’s Distance (EMD) \cite{shiGazeEMDDetectingVisual2021} algorithms. 
When the scene is not completely structured some degree of scene reconstruction is required, accomplished via RGB-D cameras \cite{wangFreeView3DGazeGuided2018, shaftiGazebasedContextawareRobotic2019, mcmullenDemonstrationSemiAutonomousHybrid2014}, stereovision \cite{cioProofConceptAssistive2019}, and machine vision techniques \cite{shiGazeEMDDetectingVisual2021}. 
In this approach, communicating intention to the system is the core challenge. 
Actions can be made context-aware, with predefined actions occurring based on looking at specific items \cite{wangFreeView3DGazeGuided2018,mcmullenDemonstrationSemiAutonomousHybrid2014} or items combined with a list of action verbs \cite{shaftiGazebasedContextawareRobotic2019} to produce the desired actions. 
However, these approaches are still limited to the number of items or grasping strategies known by the robot.
Li et al. \cite{li3DGazeBasedRoboticGrasping2017} describes an approach to additionally specify entry strategies for the robot arm, where the user identifies different points of the target item to estimate a target location and pose, and this estimation is used to define a target path for the robot. 
However, the approach is still sensitive to gaze-tracking errors, resulting in big differences in the entry angle for a small error in gaze position. 
While these automatic approaches relieve pressure from the user, it is important to note that these are limited to a set number of items, verbs and structured or calibrated environments \cite{li3DGazeBasedRoboticGrasping2017} and on a wider sense by the current limitations of robot manipulation.
Solutions tend to be computationally expensive, requiring multiple data streams for scene reconstruction and complex path planning.
Current approaches for robotic grasping and interaction are not at the readiness level needed to reliably support vulnerable people in their everyday life activities, thus, the ability to still take control of the robot to accomplish new tasks is a crucial complement to autonomous behaviour and must be developed in parallel.  
This points towards \textit{Hybrid control methods} where the user can take over and directly control the robot in edge cases \cite{wangFreeView3DGazeGuided2018}.



\subsection{Embedded Interface Approaches}


We categorize "embedded interface" as those approaches in which the interface lies on the robot itself, rather than on an external device.  
Wang et al. \cite{wangFreeView3DGazeGuided2018}, Alsharif \cite{alsharifGazeGestureBasedHuman2016} and Rico et al. \cite{ricoWAMArmModelling2014} show examples of this strategy, where the user can control the robot end-effector in a plane orthogonal to the user's point of view, looking over 2D interaction zones / dynamic commanding areas defined \textit{relative} to the end effector. 
If the user looks at a defined space relative to the effector, the robot will move in that direction, updating the zones relative to their changing position, allowing the user to issue continuous commands while looking at the robot. 
In Wang et al. \cite{wangFreeView3DGazeGuided2018} case the depth can be controlled by closing either eye, and in Alsharif et al. \cite{alsharifGazeGestureBasedHuman2016} rotation is achieved by eye-gesture controlled mode switching, requiring the user to memorize command chains.
However, it is pointed out that the inclusion of these blind interaction areas can increase the probability of erroneous commands, uncertainty, and cognitive load, making it less desirable to disabled users \cite{alsharifGazeGestureBasedHuman2016}.  
The interaction areas also remain invisible to the user during operation and do not react to changes in end effector depth, which can introduce significant ambiguity and prevent safe or fast operation. 
Movement beyond 2DOF via gaze also requires activating separate modes or using multimodal commands, burdening the user.    
The approach remains under-explored, but it is reported that embedding the interface onto the robot directly is easy to use, very intuitive \cite{wangFreeView3DGazeGuided2018}, leads to low workloads \cite{alsharifGazeGestureBasedHuman2016} and keeps attention on the end-effector at all times, avoiding the strain from context switching \cite{gabbardEffectsARDisplay2019}.







\section{Diegetic User Interface}
\label{s:DiegeticUserInterfaces}

\begin{figure}[tb]
    \centering
    \subfloat[\label{fig:UserAndbutton}User with eye-tracking glasses interacting with a button.]{\includegraphics[width=0.40\textwidth]{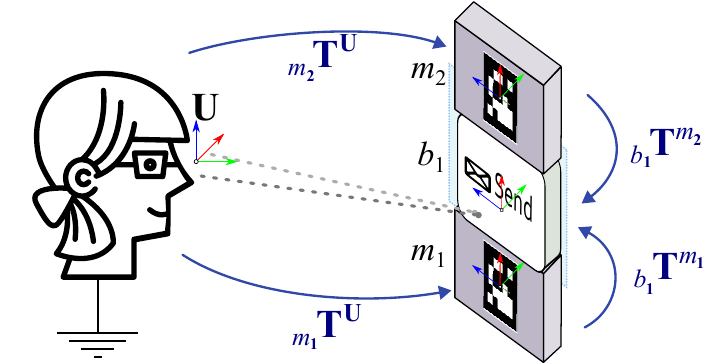}}
    \centering
    \hfill
    \subfloat[\label{fig:ButtonBasics}Diegetic Button elements. Buttons $\mathbf{B}_{i,j}$ remains fixed to its parent fiducial marker $\mathbf{M}_j$ during operation, with the transformation $_{\mathbf{B}_{i,j}}T^{\mathbf{M}_j}$ known. ]{\includegraphics[width=0.40\textwidth]{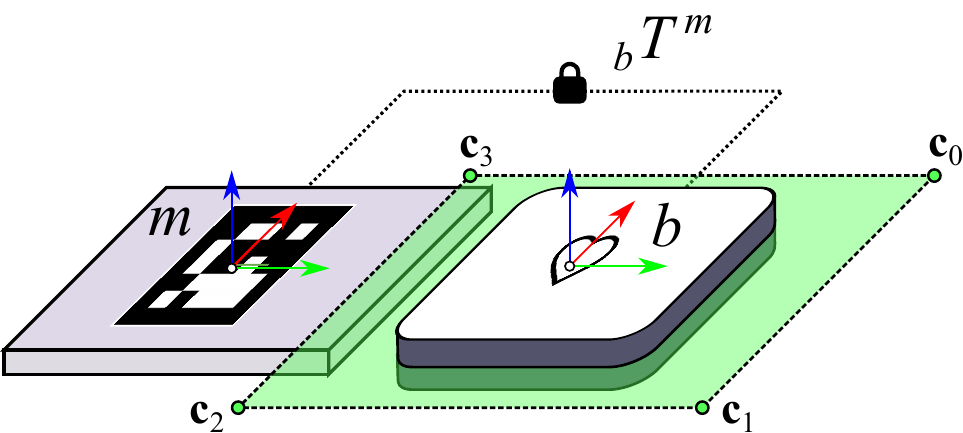}}
    \caption{Concept for Diegetic Interface system.}
\label{fig:DiegeticSystemConcept}
\end{figure}

\begin{figure*}[th]
    \centering
    \includegraphics[width=0.98\textwidth]{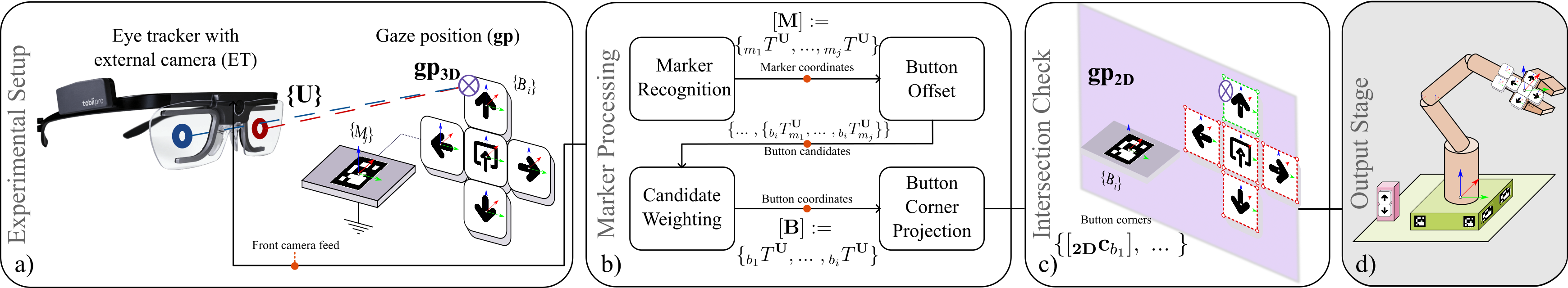}
    \caption{Camera feed and eye-gaze processing pipeline for Diegetic User Interfaces, showing from left to right 
    a) The setup with the head tracker and the diegetic buttons where the scene and point is gaze are captured.
    b) The stages of the pipeline including markers recognition, button area calculation, candidate weighting and projection back to 2D.
    c) Intersection check followed by debouncing of the signal.
    d   ) Output stage, applied to the robot.}
    \label{fig:ImageProcessing}
\end{figure*}

To alleviate the limitations presented by GUI-based, end-point, and item-based control approaches, we propose a new hybrid approach in which monitors, head-mounted output devices or projection systems are substituted with a \textit{Diegetic Graphical User Interface} (D-GUI), a screenless, computationally inexpensive GUI embedded in the real world or robot. 
The system works similarly to existing GUI approaches, in which the gaze acts as a cursor, selecting behaviour from buttons (Figure \ref{fig:UserAndbutton}), detected with a commercial wearable eye-tracker with an outward-facing camera, matching where the user is looking in the camera feed.
The buttons are localized by pairing them with fiducial markers, which can be easily found by machine vision techniques.
These \textit{Diegetic Buttons} have a fixed position relative to the markers, with the transformation between them known by the system beforehand (Figure \ref{fig:ButtonBasics}).  
The user can interact with this \textit{diegetic interface}, placed physically around the person or onto the robot itself, and can be reconfigured in real-time to their liking. 
The system can understand when the user is looking at the buttons and then trigger different actions after a specified time (dwell time) has passed. 
Diegetic is defined in works of fiction as elements that exist in the character's world and are perceptible to them \cite{DiegeticDef}.
Thus, our diegetic interface exists concretely in the real world and is perceived naturally by the user and people surrounding them, as opposed to on-screen or augmented reality interfaces that might rely on virtual elements to enable perception. 

This section describes the process of finding a button's position and rotation based on multiple marker candidates, evaluating interaction with eye-gaze, and filtering strategy for control, summarised in Figure \ref{fig:ImageProcessing}. 
We adopt the convention ${}_B \mathbf{P} = {}_B T^A \; {}_A \mathbf{P}$, where ${}_A \mathbf{P}$ and  ${}_B \mathbf{P}$ are points relative to the frames A and B respectively, and  ${}_B T^A$ is the transformation of frame $\mathbf{A}$ to frame $\mathbf{B}$. 
Similarly ${}_B R^A$ is the rotation of frame $\mathbf{A}$ to frame $\mathbf{B}$.

\subsection{Diegetic Button Detection}
\label{ss:DiegeticButtonDetection}


The interactive components are the \textit{Diegetic buttons}, comprised of one or more interactive user interface (U.I.) elements and a fiducial marker used for localization. 
A fiducial marker is a tag with highly distinguishable characteristics that can be easily recognized by the imaging system. 
Fiducial markers can be used for pose estimation, where their transformation relative to the camera can be derived easily by analysing its dimensions and rotation \cite{kalaitzakisFiducialMarkersPose2021}
ArUco tags are a particular type of these \cite{ArUco-garrido-juradoAutomaticGenerationDetection2014,kalaitzakisFiducialMarkersPose2021}, widely used in robotics applications. 
We define the button positions fixed relative to multiple of these ArUco markers so that if a marker is visible the button transformation and bounding box can be easily found. 
Note that with this approach the calculation of the transformation between the world reference frame $\mathbf{W}$ or the user frame $\mathbf{U}$ is not necessary.

From the eye-tracking glasses, a video feed is received from the front-facing camera. 
From each individual image of the video feed, the OpenCV implementation of ArUco marker recognition \cite{opencv_library} is used to extract the transformation from a marker $m$ to the camera/user frame of reference $\mathbf{U}$ obtaining the homogeneous transformation matrix $_{\mathbf{m}}T^U$.
This is done for all the fiducials in the image, obtaining a collection of marker transformations 
$[ \mathbf{M} ] := 
\{ 
{}_{\mathbf{m_1}}T^{\mathbf{U}} , 
{}_{\mathbf{m_2}}T^{\mathbf{U}} , 
\dots, 
{}_{\mathbf{m_j}}T^{\mathbf{U}} 
\}$.  


To find the button positions and properties, the position and rotation of each button $\mathbf{b}_i$ relative to a marker $\mathbf{m}_j$ is defined prior via the transformation $_{\mathbf{b}_i}T^{\mathbf{m}_j}$, forming a parent-child relationship. 
To simplify notation, the following calculations omit the marker and button sub-indexes, referring to a general individual marker as $\mathbf{m}$ and button as $\mathbf{b}$ respectively and their transformation as $_\mathbf{b}T^{m}$.
This transformation is measured while designing the elements and configured before using the device, with a local position (${}_{\mathbf{m}}\mathbf{P}_\mathbf{b}$) and local rotation (${}_\mathbf{b}R^{\mathbf{m}}$) from the marker:

\begin{align}
    {}_\mathbf{b}T^{\mathbf{m}} 
    = 
    \begin{bmatrix} {}_{ \mathbf{b}} R^{\mathbf{m}}  &  {}_{\mathbf{m}}\mathbf{P}_{ \mathbf{b}} \\ o & 1 \end{bmatrix} 
\end{align}

Thus a candidate for the transformation from the User frame $\mathbf{U}$ to a button $\mathbf{b}$ via a specific marker $\mathbf{m}$ can then be found by the composition:

\begin{align}
{}_\mathbf{b}T^\mathbf{U}_\mathbf{m} = 
{}_\mathbf{b}T^{\mathbf{m}} \; 
{}_{\mathbf{m}}T^\mathbf{U} 
\end{align}


However, as the marker recognition can be distorted by camera quality, errors in camera parameters, illumination conditions, and other factors, the derivation of the transformation from user to marker ${}_{\mathbf{m}}T^\mathbf{U}$ is not error-less.
This error propagates and increases with the distance from the button to marker ${}_\mathbf{b}T^{\mathbf{m}}$, resulting in more inaccurate estimation as the distance from the marker increases. 
To improve the estimation accuracy of the button pose, and in case of marker occlusion, we use multiple markers to calculate the position of a single button when these are available to increase robustness. 
Multiple candidates for position and rotation are pooled together to obtain a better estimate. 

To obtain the final estimated position for single button ${}_{\mathbf{U}} \mathbf{P}_\mathbf{b} $, we opt for a weighted average of all positions, with higher influence given to markers closer to the button.
Weights $\omega_{\mathbf{m},\mathbf{b}}$ are first calculated from a button $b$ to each of its parent markers $\mathbf{m}$, defined as power-n inversely proportional to the norm distance to each parent marker $b$:

\begin{align}
\omega_{\mathbf{m},\mathbf{b}} = \dfrac{1}{ | 
{}_{\mathbf{m}}\mathbf{P}_{\mathbf{b}}
|^n }
\end{align}

A value of $n=2$ was set in our implementation, found empirically as a trade-off between the influence of distant and close markers. 
The final position candidate ${}_{\mathbf{U}} \mathbf{P}_\mathbf{b} $ is thus given by the weighted average: 

\begin{align}
{}_{\mathbf{U}} \mathbf{P}_\mathbf{b}  = 
\dfrac{ \sum_{\mathbf{m}}  ( \;  
{}_{\mathbf{U}}T^{\mathbf{m}} \; 
{}_{\mathbf{m}}\mathbf{P'}_{\mathbf{b}} \; 
\omega_{\mathbf{m},\mathbf{b}} \; ) 
}{
\sum_{\mathbf{m}} \; ( \omega_{\mathbf{m},\mathbf{b}} ) } 
\end{align}

Where ${}_{\mathbf{m}}\mathbf{P'}_{ b} $ is the dimension corrected $4 \times 1$ position vector:

\begin{align}
     {}_{\mathbf{m}}\mathbf{P'}_{\mathbf{b}} 
     =
    \begin{bmatrix}   
        {}_{\mathbf{m}}\mathbf{P}_{\mathbf{b}} 
        \\ 0 
    \end{bmatrix} 
\end{align}


Final rotation is found using the same method, obtaining the weighted intermediate angle between the candidates' rotation. 
This has been implemented as described by Landis et al \cite{markleyAveragingQuaternions2007}, by averaging the quaternion representation ${}_\mathbf{b} \overline{  q } ^{\mathbf{U}}$ of the final rotations $ {}_\mathbf{b}R^{\mathbf{U}} =  {}_\mathbf{b}R^{\mathbf{m}} \; {}_{\mathbf{m}}R^{\mathbf{U}} $ using the $\omega_{\mathbf{m},\mathbf{b}}$ weights, controlling for the uniqueness of representation and normalizing. 
The rotation is then returned to matrix representation to obtain ${}_\mathbf{b}R^{\mathbf{U}}$. 

Finally, rotation and position candidates can be combined to obtain the homogeneous transformation candidates from the user to each button:

\begin{align}
    {}_\mathbf{b}T^{\mathbf{U}} = \begin{bmatrix} {}_\mathbf{b}R^{\mathbf{U}} &  {}_{\mathbf{U}} \mathbf{P}_\mathbf{b} \\ o & 1 \end{bmatrix} 
    \label{eq:transform}
\end{align}

Resulting in a list $\mathbf{B}$ of the final user-to-button transformations  
$[ \mathbf{B} ] := 
\{ 
{}_{\mathbf{b}_1}T^{\mathbf{U}} , 
{}_{\mathbf{b}_2}T^{\mathbf{U}} , 
\dots, 
{}_{\mathbf{b}_i}T^{\mathbf{U}} 
\}$.

\subsection{Bounding Box Projection}
\label{ss:BoundingBoxProjection}

To evaluate whether the user is looking at the button, an interaction zone is defined as a rectangular zone relative to the button centre for each button in $[\mathbf{B}]$, projected into 2D camera coordinates and evaluated for overlap with the 2D gaze point of the user.
Each interaction area is comprised of four corner points: $[{}_\mathbf{b} \mathbf{c} ] 
:= \{ {}_\mathbf{b} \mathbf{c}_{1} , \;
\cdots ,
{}_\mathbf{b} \mathbf{c}_{4} \}$, shown in Figure \ref{fig:ButtonBasics}, defined relative to their respective button $\mathbf{b}$ center. 
The transform referenced from the user frame can thus be found using:

\begin{align}
    [ 
    {}_{\mathbf{U}} \mathbf{c}
    ]
    &=
    {}_{U}T^\mathbf{b} \; 
    [ 
    {}_{\mathbf{b}} \mathbf{c}
    ]
\end{align}

The eye tracker returns gaze position values expressed in the 2D pixel coordinate frame of the camera. 
We project the button positions from the 3D global frame to the camera frame to enable direct comparison.
The boundaries are then projected to 2D camera coordinates using the camera pinhole model and the camera calibration parameters \cite{opencv_CameraCalibrationand3DReconstruction}.
The 3D points $[ {}_{\mathbf{U}} \mathbf{c} ]$ relative to camera frame $\mathbf{U}$ can be transformed into 2D pixel coordinates $[ {}_{\mathbf{2D}} \mathbf{c} ]$ using the constant camera intrinsic matrix $K$, relative to the camera internal $\mathbf{2D}$ frame of reference:


\begin{align}
    [ {}_{\mathbf{2D}} \mathbf{c} ]
    = K \;
    [ {}_{\mathbf{U}} \mathbf{c} ]
\end{align}

Where the matrix $K$ represents the intrinsic characteristics of the camera, here the Tobii Pro Glasses 2: 

\begin{align}
\label{eq:k}
    K = 
    \begin{bmatrix}
    f_x & 0 & c_x & 0 \\
    0 & f_y & c_y & 0 \\
    0 & 0 & 1 & 0
    \end{bmatrix}
\end{align}


Where $K$ is composed of the focal lengths $f_x$ and $f_y$ in pixels and the principal point close to the image centre  $c_x, c_y$, found experimentally \cite{opencv_CameraCalibrationand3DReconstruction}.
Thus, the final position for each corner is described by:

\begin{align}
    [ {}_{\mathbf{2D}} \mathbf{c} ]     
    &= K \;  
    {}_{U}T^{\mathbf{b}} \; 
    [ 
    {}_{\mathbf{b}} \mathbf{c} 
    ] 
\end{align}

The four corners are then used to define a closed polygon area $A_b$. 
If the 2D gaze position $\mathbf{gp_{2D}}$ is contained inside this area it is considered that the user is looking and interacting with a button, and the input processing stage begins.

\subsection{Input Processing: Dwell time}
\label{ss:InputProcessing}


Inputs are filtered and debounced following a dwell-time strategy, preventing accidental inputs by requiring the user to fixate for a short time. 
Every time step for each active input in the list, an activation parameter $a_b(t)$ is updated.
If a button is active its activation parameter is increased by $\Delta t / T $ and decreased by the same amount if inactive, where $\Delta t = t' - t$ is the change in time since the last iteration and $T$ is the activation period for the buttons corresponding to the dwell time.
Values are limited to $0 \leq a_\mathbf{b} \leq 1$, with 0 corresponding to the minimum activation value and 1 to the maximum.
The process is analogous to a rolling average, with the number of samples given by $ T / \Delta t $.
The final activation parameter $a_{\mathbf{b}}(t')$ is thus defined as: 

\begin{align}
    a_{\mathbf{b}}(t') = \left\{ 
    \begin{array}{ll}
        a_{\mathbf{b}}(t)  + \Delta t / T &  \mbox{if $\textbf{gp}$ in $\{ {}_{\mathbf{2D}} \mathbf{c}_{\mathbf{b}} \}$ and $a_{\mathbf{b}}(t) \leq 1\}$}\\
        a_{\mathbf{b}}(t)  - \Delta t / T & \mbox{if $\textbf{gp}$ out $\{ {}_{\mathbf{2D}} \mathbf{c}_{\mathbf{b}} \}$ and $a_{\mathbf{b}}(t) \geq 0\}$}\\
        a_{\mathbf{b}}(t) & \mbox{otherwise}
        \end{array} \right. 
\end{align}


Following this, a Schmitt trigger is applied. 
Two parameters $a_{\text{on}}$ and $a_{\text{off}}$ are used to define input behaviour. 
If $a_{\mathbf{b}} \geq a_{\text{on}}$ the buttons activation flag is set to an \textit{active} state, and if $a_{\mathbf{b}} \leq a_{\text{off}}$ the buttons is considered \textit{inactive}: 

\begin{align}
    \text{status}_i(t) = \left\{ 
    \begin{array}{ll}
        \mbox{active} &  \mbox{if $a_{\mathbf{b}}(t) > a_{\text{on}}$}\\
        \mbox{inactive} & \mbox{if $a_{\mathbf{b}}(t) < a_{\text{off}}$}
        \end{array} \right. 
\end{align}

The collection of buttons with their IDs, activation values $a_b$ and activation flags is then sent to the controller node.   
At the end of each cycle, the active buttons are stored in a list of active inputs, and all buttons and their activation values are packet and sent. 
Actions associated with a given button ID can then be triggered.

\begin{figure}[t]
    \centering
    \includegraphics[width=0.35\textwidth]{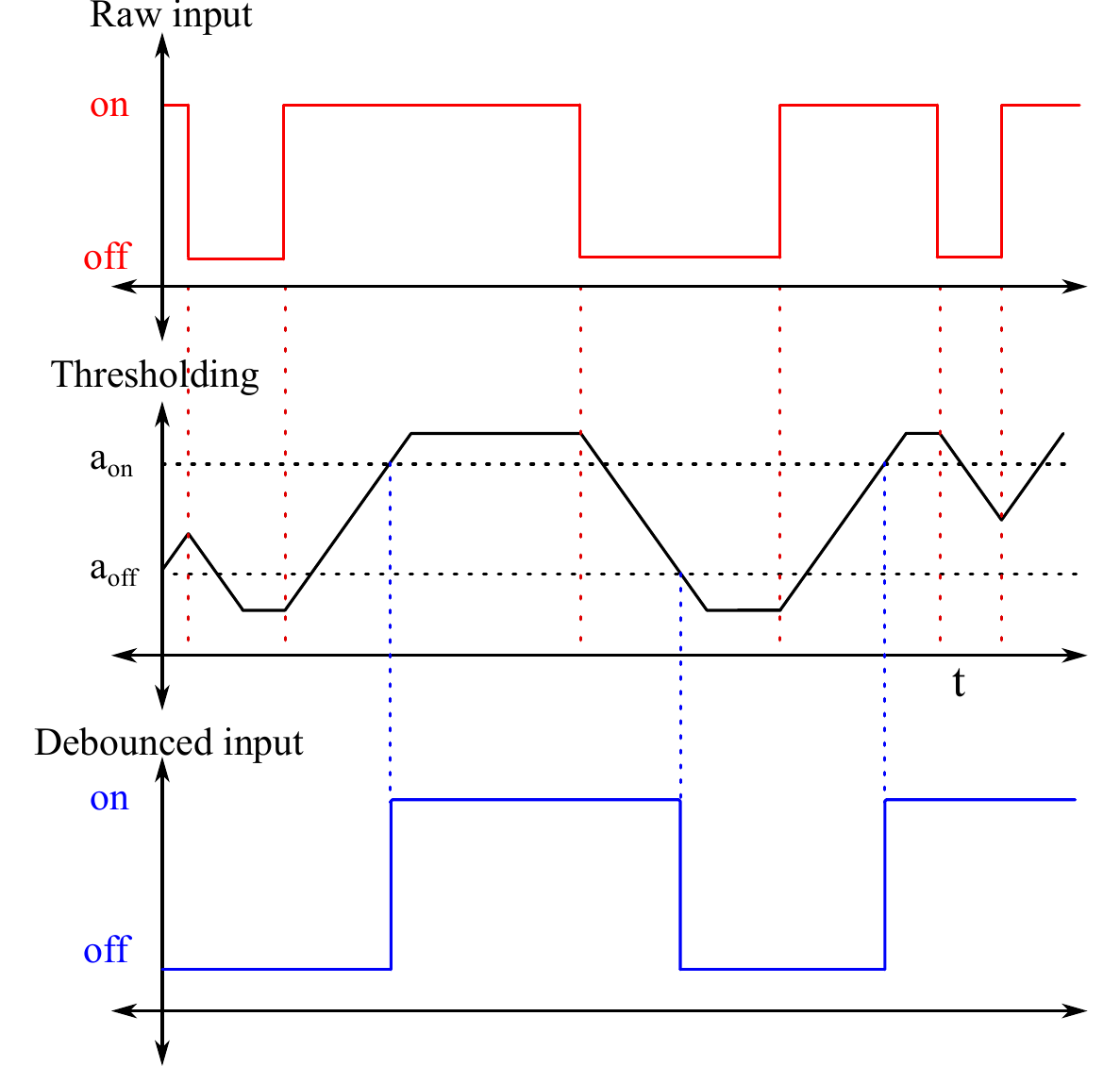}
    \caption{Input signal processing sample, showing raw input from intersection check (top), rolling average with limits and thresholding marks (middle), and final debounced signal with a Schmitt trigger (bottom).
    The parameters $a_{\text{on}}$ and $a_{\text{off}}$ define the activation of the end signal. }
    \label{fig:InputDebouncing}
\end{figure}

The control of the robot's actions is handled differently depending on if the action is continuous or discrete. 
Discrete actions such as opening and closing the end-effector gripper are associated with single buttons and are triggered when their active state is reached. 
A dwell-time window of $T_{d}=1000\text{ ms}$ is used, consistent with common dwell times for screen-based interactions (between 500 ms and 1000 ms) \cite{GazebasedTeleprosthetic_Dziemian}. 
Discrete actions have activation values $a_{\text{on}}= a_{\text{off}}= 0.8$ close to 1 to reduce triggering accidentally while providing a small filtering window. 
Figure \ref{fig:InputDebouncing} illustrates the input processing behaviour. 


For continuous actions, in this case the end-effector movement, we take a modified approach. 
Dwell time selection needs to be large enough to prevent unintentional interactions, approximately larger than $300$ms \cite{shiGazeEMDDetectingVisual2021}.
However, repeated small dwell time waits still accumulate and slow down the interaction, resulting in a cognitive burden \cite{fujiiGazeGestureBased2018}. 
To resolve this, we propose that discrete actions benefit from a traditional dwell time approach, while continuous actions benefit from activation ramps. 
We separate end-effector movement into two phases: 
(1) A constant acceleration phase during a set period $T_c$ towards, 
(2) a maximum target speed $v_{\text{ref}}$. 
These stages are analogous to a dwell time window, and discrete activation respectively. 
The user receives close-to-immediate feedback in the form of subtle end-effector movement while minimizing the effect of accidental inputs.
Activation values were set to $a_{\text{on}}=0.4$, $a_{\text{off}}=0.2$, set empirically to provide a small activation value, with an acceleration window of $T_c = 300$ms. 
The velocity is thus a function of the activation coefficients $a_b$ of the active buttons and proportional to the user interaction time. 
The end-effector thus starts moving. when the user has looked at it for $T_c \cdot a_{\text{on}} = 120$ms at 40\% of its maximum speed and accelerates to its maximum for $160$ms. 


This approach also allows small delicate movements via repeated saccades, while providing immediate feedback to the user in the way of small movements. 
A constant acceleration is used here, with more complex profiles left as the target of future studies.


\subsection{Robot D-GUI and Control Strategy}

\begin{figure}[tb]
    \centering
    \subfloat[]{\includegraphics[width=0.23\textwidth]{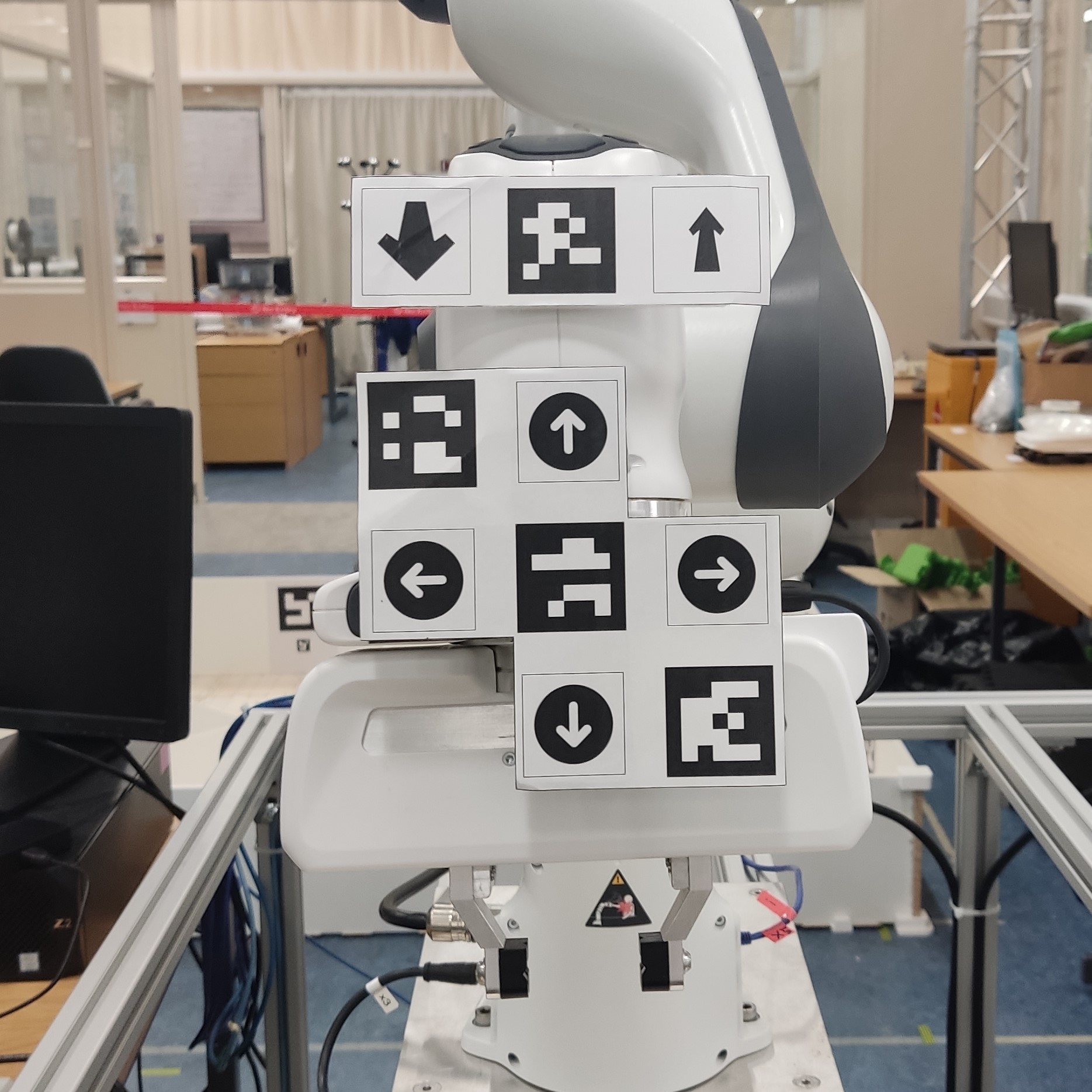}
    \label{fig:Robot3DOFControl}
    }
    \centering
    \hfill
    \subfloat[]{\includegraphics[width=0.23\textwidth]{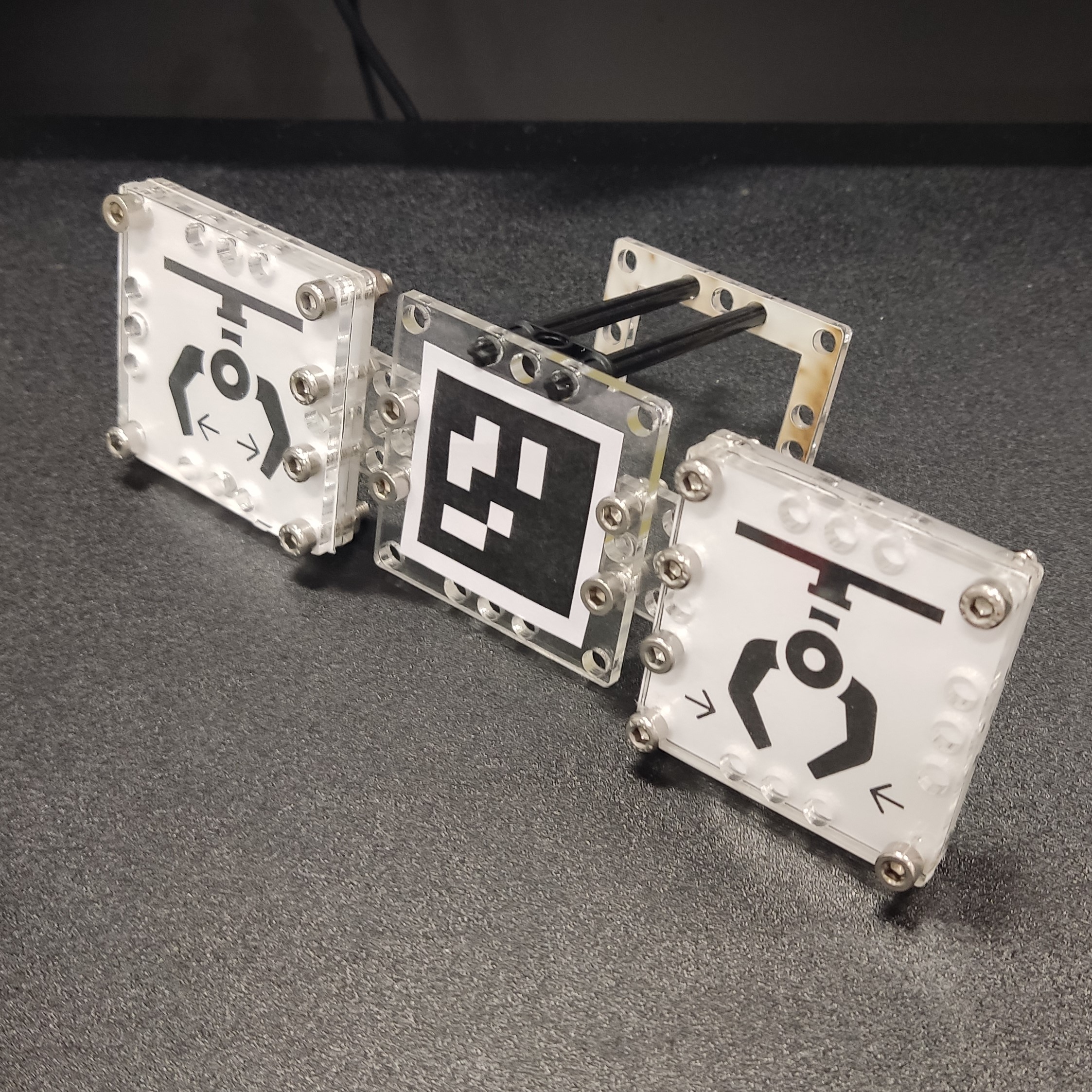}
    \label{fig:RobotGrabcontrol}
    }
    \caption{User interface setup. a) Directional arrows for 3DOF control. Top arrows control depth, bottom ones control planar movement relative to the user.
    b) External controls for opening/closing the gripper.}
\label{fig:RobotUI}
\end{figure}

The proposed diegetic user interface for the robot is shown in Figure \ref{fig:RobotUI}.
The robot has a removable marker on the front of the end-effector with the interactive buttons, used to control Cartesian position (Figure \ref{fig:Robot3DOFControl}). 
The centre arrow moves the robot aligned with its world coordinates orthogonally relative to the user. 
The arrows on the bar on top control depth, the left arrow brings closer to the user and the right arrow pushes away. 
A separate set of buttons controlling the closing or opening of the end-effector is placed on the table next to the cubes (Figure \ref{fig:RobotGrabcontrol}). 
As the experiment focuses on 3DOF cartesian control, no control over rotation will be given to the user and the arm rotation is locked in place.

Control is achieved by defining antagonistic inputs (i.e. moving up against moving down) and obtaining the difference in activation values of buttons in an active state.  
To obtain the final end-effector velocity $v_{ee}$, the difference between inputs is multiplied by a target speed in each of the respective axes: 

\begin{align}
v_{ee} =  
    \begin{bmatrix}
    v_{\text{ref x}} \\
    v_{\text{ref y}} \\
    v_{\text{ref z}} 
    \end{bmatrix} 
    \begin{bmatrix}
    ( a_{\text{left}} - a_{\text{right}} )\\
    ( a_{\text{forward}} - a_{\text{backward}} )\\
    ( a_{\text{up}} - a_{\text{down}} )
    \end{bmatrix}
    \text{m/s}
\end{align}

The velocity is then applied to the end-effector cartesian speed using the Moveit library. 
The end-effector speed of the robot was set to $v_{\text{ref}}=[ 0.1 , 0.06 , 0.08 ]\; \text{m/s}$, and the activation values to produce an equivalent acceleration period of $T_{c}=400$ ms, set empirically.
For the velocity values depth movement was set at a lower speed to compensate for the increased difficulty in judging depth compared to the other directions. 
The acceleration period was set close to the minimum recommended by Shi et al. \cite{shiGazeEMDDetectingVisual2021}.

As a safety measure, the controller is limited to a maximum force of 30N for the end-effector, which if encountered will immediately trigger an emergency stop.   
Robot speed was limited to 0.6 m/s and angular joint speed to 1 rad/s.
A virtual cage has also been defined surrounding the robot, which will impede robot movement if attempted to cross.

\section{Implementation}
\label{s:RobotControl}



\subsection{Hardware}

The Tobii Pro Glasses 2 (Tobii AB, Sweden) were chosen as a head-mounted gaze-tracking detection interface \cite{TobiiProGlasses2}. 
The device provides real-time streaming of video and eye data at a 50Hz frame rate.
3D and 2D gaze positions and an outside camera view can be obtained in real-time for processing.
The glasses are also portable, connected to the computer via WiFi allowing untethered use. 
The Tobii Pro Glasses 2 were selected due to their availability and natural look, similar to commonly used glasses, which has been described as a priority and factor of acceptance by people with tetraplegia \cite{AssessmentofBrainMachineInterfacesfromthePerspectiveofPeoplewithParalysis_Blabe}.

For the robotic arm, a Panda arm (Franka Emika, Germany) \cite{PandaArm} was selected. 
While this arm is mostly used for industrial applications, here it was chosen as the test platform to demonstrate the capabilities of the system due to its ease of integration with ROS environments, compliance, and multiple safety features.
A PC running Ubuntu 20 with real-time kernel was used to run the software.
The computer is directly connected to the Franka Arm via Ethernet and to the Tobii Pro Glasses 2 via WiFi. 
Robot operations run at 200Hz.

To minimize weight, markers and button pictures were printed on paper and attached to hard cardstock. 
The markers and buttons attached to the Franka Arm are held with velcro tape, allowing repositioning buttons and trying different arrangements. 
External buttons are held in a small acrylic cut platform. 

\subsection{Software}

\begin{figure}[tb]
\centering
\includegraphics[width=0.42\textwidth]{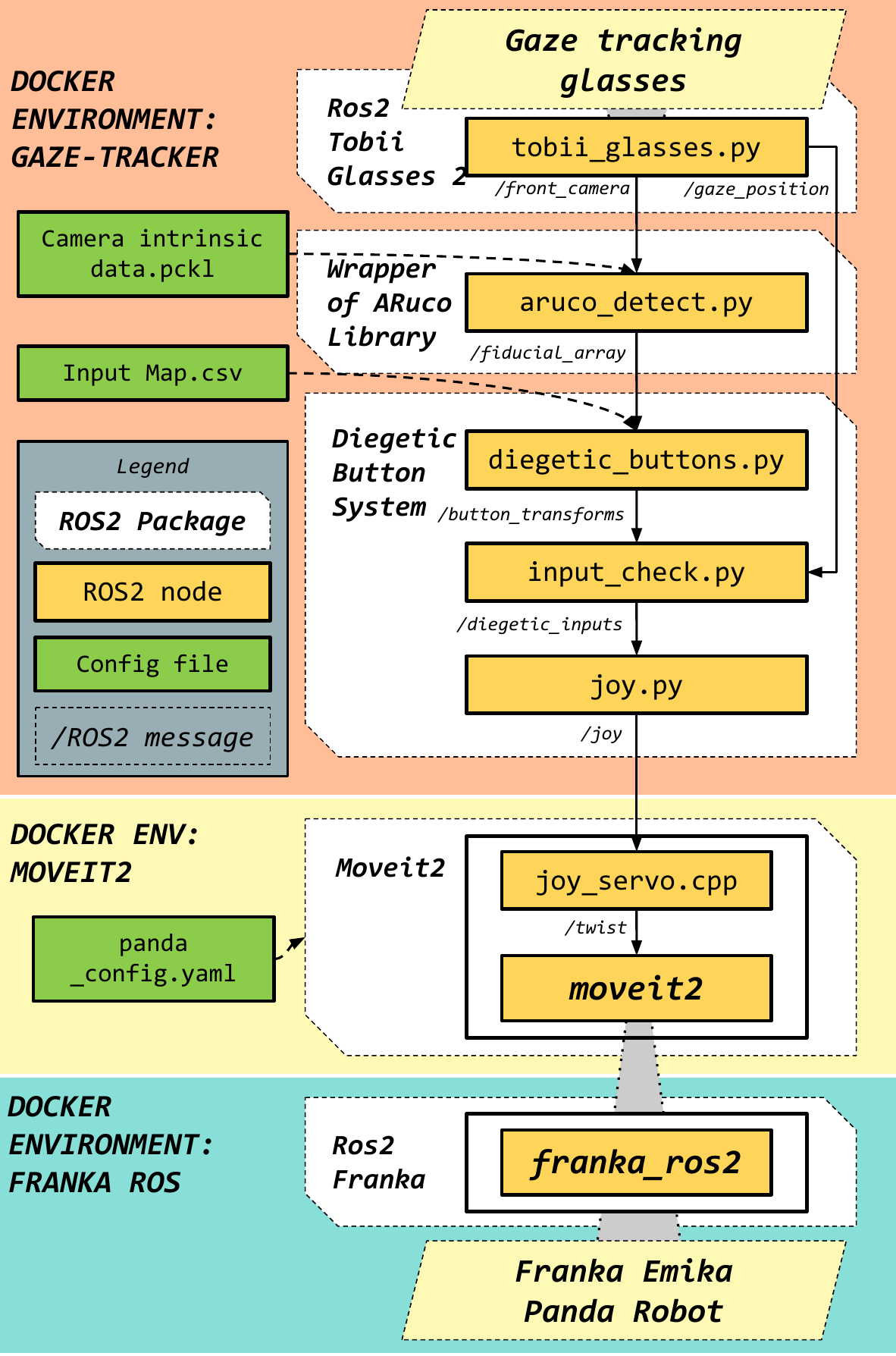}
\caption{Diagram for the system software architecture.}
\label{fig:SoftwareArchitecture}
\end{figure}

The system has been implemented using ROS 2 Foxy (Robot Operative System version 2) \cite{ROS2} as middleware, with the individual components implemented as a collection of nodes in different environments. 
It was found during development that some of the libraries used have contradictory requirements.
To manage dependencies consistently, ROS 2 is running on \textit{Docker containers} defining the specific setups. 
Thus, the Docker virtualization tool was used to define distinct environments with the respective software package requisites, while allowing the ROS nodes to communicate.
The general system architecture has been summarized in Figure \ref{fig:SoftwareArchitecture}, showing the three Docker containers for (a) the Tobii Pro Glasses 2 connection and visual processing elements, (b) the MoveIt2 velocity-based controller and robot configuration, and (c) the FrankaROS 2 environment for communication with the robot.


To enable the connection between the Tobii Pro Glasses 2, we used our re-configurable ROS 2 node and Docker environment, previously presented in \cite{NunezSardinhaOpensourceTobiiProGlasses2withROS2_2022}.
The package interfaces with the glasses within ROS 2 and provides a single topic with the camera feed, gaze position, and other sensor data.
To process the data from the glasses, two new software packages were created: the first, \textit{ARuco detect} consists of a simple wrapper for the Open-cv ArUco fiducial pose estimation algorithm \cite{ArUco-garrido-juradoAutomaticGenerationDetection2014} for ROS2.
The package receives the camera feed of the glasses, processing and sending a collection of fiducial marker positions as outputs. 
This requires the Tobii Pro Glasses 2 camera intrinsic parameters (Equation \ref{eq:k}), which were found experimentally and stored in an external file.

The second package performs the diegetic buttons tasks, and it's comprised of 3 nodes.
The node \textit{Diegetic Buttons} receives the fiducial markers transform information and calculates the position of the buttons with their bounding area, as described in Section \ref{ss:DiegeticButtonDetection}. 
The marker-button relationships are defined in a CSV file that is accessed before run-time. 
The \textit{Input check} node evaluates the intersection between eye gaze and the button bounding area (Section \ref{ss:BoundingBoxProjection}) and constructs a list of buttons with their associated action, activation parameter and current state. 
The \textit{joy} node translates the input information into joystick commands, widely used across ROS 2 distributions. 
This allows the system to emulate a general joystick controller and eases development and integration with other systems using the same interface.

The control was implemented using the MoveIt Motion Planning Framework for ROS 2 \cite{Moveit2}.
A general motion planning and control framework for robots.
These packages simplify higher-level control of the system.
The \textit{servo package} is used as an intermediary to send commands and is used as the base to implement the current controller for the robot.
The interface receives target changes in speed and rotation of the end-effector and performs inverse kinematic calculations to reach these targets in real-time via the inverse Jacobian.  
The robot will attempt to match this speed unless a force higher than a set threshold is perceived, in our case set to 30N.
The \textit{joy servo} node acts as an intermediary between the joy message from the glasses package and MoveIt, translating joy commands into the target Cartesian velocity and gripper instructions to the servo system.
It also defines safe zones forming a box around the interaction area, preventing the robot from accidentally being driven outside of the working space boundary. 
Currently, control is constrained to 3DOF cartesian directions and a controllable gripper claw, with locked rotation.

The used Docker environment with the contents of Moveit2, Moveit2 Tutorials and the ROS 2 Franka library has been prepared according to official installation instructions available under a GNU V3.0 license (Appendix \ref{s:OpenSourcePackages}).
Control of the \textit{Franka Emika arm} is done using the interfacing library \textit{ROS 2 franka}, which uses libfranka to connect directly to the robot using the Franka Control Interface.

\begin{figure}[thb]
    \centering
    \includegraphics[width=0.49\textwidth]{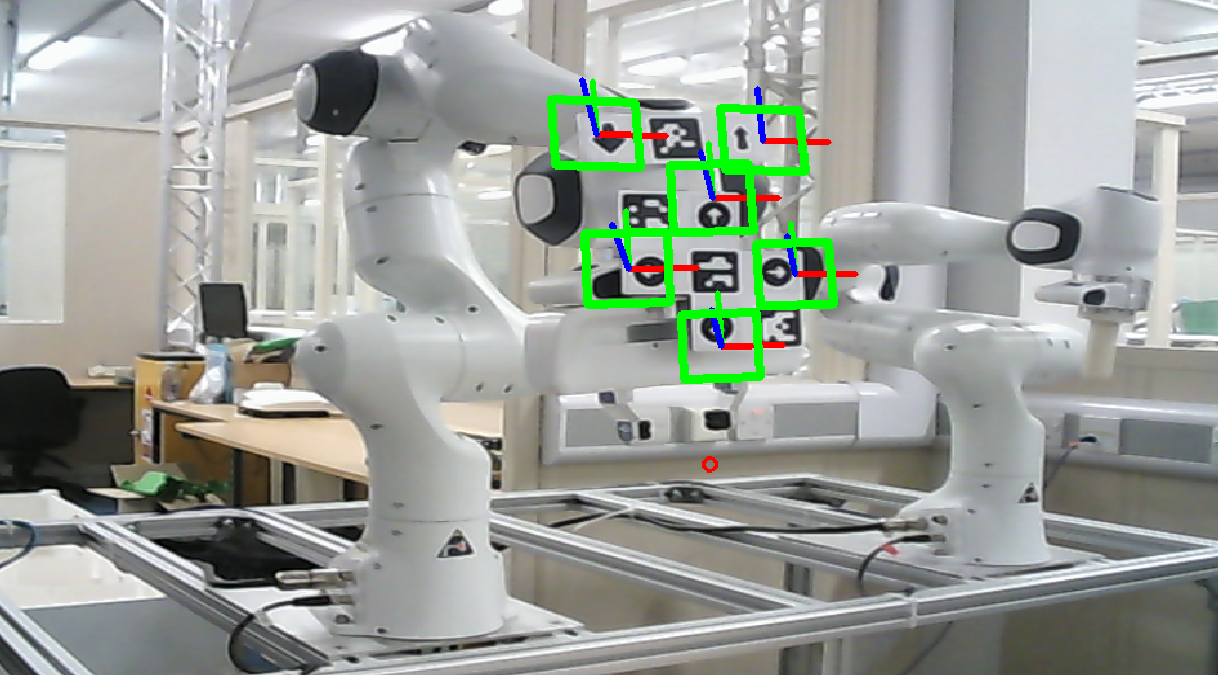}
    \caption{Robot arm with mounted interface and highlighted input overlay. To account for jitter in the user gaze and errors in calibration, interaction zones have been defined as slightly larger than the actual button size.
    Note that the user perceives the unaltered reality only and the highlights remain in software only.
    }
    \label{fig:RobotOverlay}
\end{figure}


\subsection{Preliminary Results and Performance}

The proposed interface is shown in Figure \ref{fig:RobotOverlay} with interaction zones highlighted.
The controls on the end-effector allow 3D movement with the axes relative to the robot base, and the separate controls on the table allow grasping. 
To compensate for the small size of the buttons and errors in gaze point calibration the interaction zones are defined as larger than the actual size of the buttons (56x56mm against the button's 36x36mm area, or an extra 1cm per side).




To obtain estimates of the performance of the system, time values were sampled during a minute of system activity while being controlled by a user.
Controller operation was measured to run at $49.401 (\pm 0.00726)$Hz.
The average message coming from the glasses had a size of 1.56MB for a bandwidth of 78MB/sec, composed primarily of front camera feed and by gaze information. 
Process latency was measured for every node with the processing pipeline taking  $30 ( \pm 14.17)$ms to complete, and cumulative latency is shown in Table \ref{Tab:ProcessLatency}. 
Latency values for communication with the Tobii Pro Glasses 2 were measured to be $11.96 ( \pm  23.91)$ms, with occasional spikes surpassing 200 ms depending on proximity to the host machine.  
Total latency is estimated at 42ms, making it suitable as an input device.  
The wireless communication between the eye-tracker and the system is suspected as the main operational bottleneck, and future approaches will opt for tethered connections instead.

\begin{table}[tb]
    \centering
    \caption{Cumulative latency times for our gaze-interaction implementation execution.     }
    \label{Tab:ProcessLatency}
\begin{tabular}{cc}
Process                         & Latency (ms)                \\ \hline
tobii\_glasses.py & $12 ( \pm 4.64)$  \\
aruco\_detect.py        & $23 ( \pm 11.28)$   \\
button\_transforms.py          & $26 ( \pm 13.07)$   \\
input\_check.py                         & $28 ( \pm 13.99)$  \\
joy.py (Final total)                         & $ \mathbf{30 ( \pm 14.17)}$ \\\hline
\end{tabular}
\end{table}



\begin{figure}[t]
    \centering
    \includegraphics[width=0.49\textwidth]{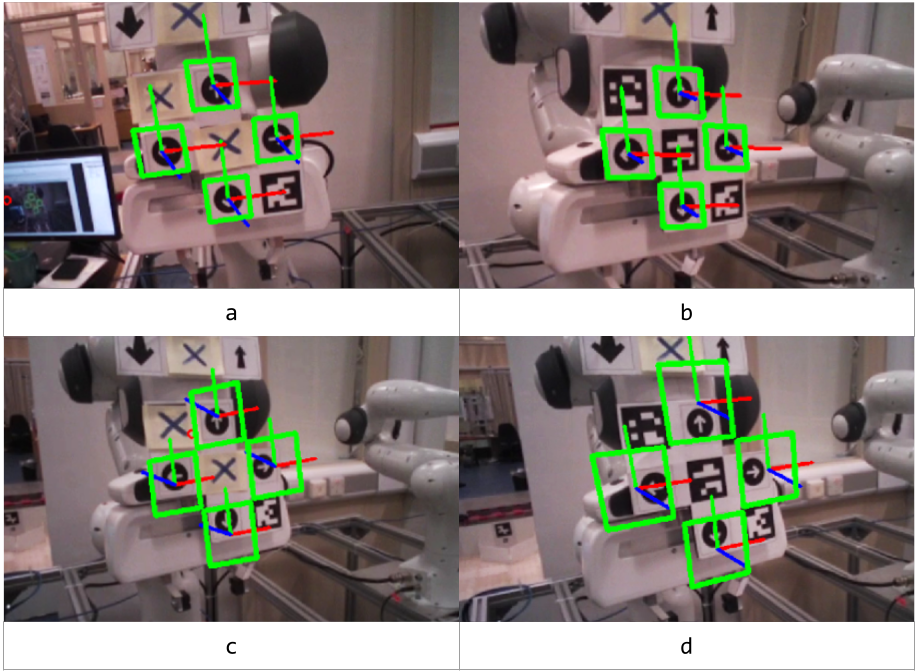}
    \caption{Highlighted interaction zones of the robot, showing areas from a different number of markers.
    Figures a) and c) show zones from a single marker in the corner, and b) and d) from the three markers in the cross.
    The top row shows default interaction zones, bottom row shows expanded zones for compensation. 
    Errors in interaction areas are magnified with fewer markers.
    }
    \label{fig:PianoOverlay}
\end{figure}

Figure \ref{fig:PianoOverlay} briefly illustrates the effect of multiple markers on buttons on a sample test interface. 
Markers were covered to illustrate the difference between relying on one or multiple markers for calculation of interaction zones. 
Images were captured with the Tobii Pro Glasses 2, and the interaction area was highlighted.
As markers are visible to the outside camera of the eye-tracking glasses, alignment between the virtual and real zones improves. 
The effect is more notable in the extended interaction zones.

\section{User Study Methods}
\label{s:Methods}

To evaluate the dexterity and accuracy of the presented system, a slightly modified version of the \textit{Yale-CMU-Berkeley (YCB) Block Pick and Place Protocol} \cite{BenchmarkingManipulationResearchUsingYCBObjectandModelSet_Calli_2015} has been performed.
The test was selected as it is recommended to evaluate precision and dexterity in robotic manipulation for everyday tasks resembling  Activities of Daily Living. 
Unlike other precision tests such as Peg Insertion Tests, the user cannot rely on the pegs being caught in a slot to complete the task and needs to precisely place the boxes in place before releasing.

In the test the user controls the robot to pick up and place 2cm blocks on a stencil, marking square target zones. 
The user receives points after placing each box correctly: one point if the block is partially inside the square and two if it is completely inside. 
No points are received if the block is placed on the unmarked  stencil (white paper area).
Points are subtracted if the user places the boxes outside the stencil: one if it is partially out, and two if it is completely.
The pick-and-place task must be performed 8 times for each mark on the stencil, after which the test finishes (for a maximum of 16 points). 
The score is thus considered proportional to the user's accuracy and dexterity.
Timestamps for pick-up and placing have been recorded to capture a potential learning period.
Due to the limited workspace and rotation constraint of the end effector, the start conditions for the test were modified: the cubes were placed next to the stencil on either side rather than 10cm away from it, and their rotation was aligned with the squares in the stencil (Figure \ref{fig:RobotSetup}). 
No other changes to the YCB protocol were made. 


\subsection{Setup}

\begin{figure}[tb]
    \centering
    \includegraphics[width=0.48\textwidth]{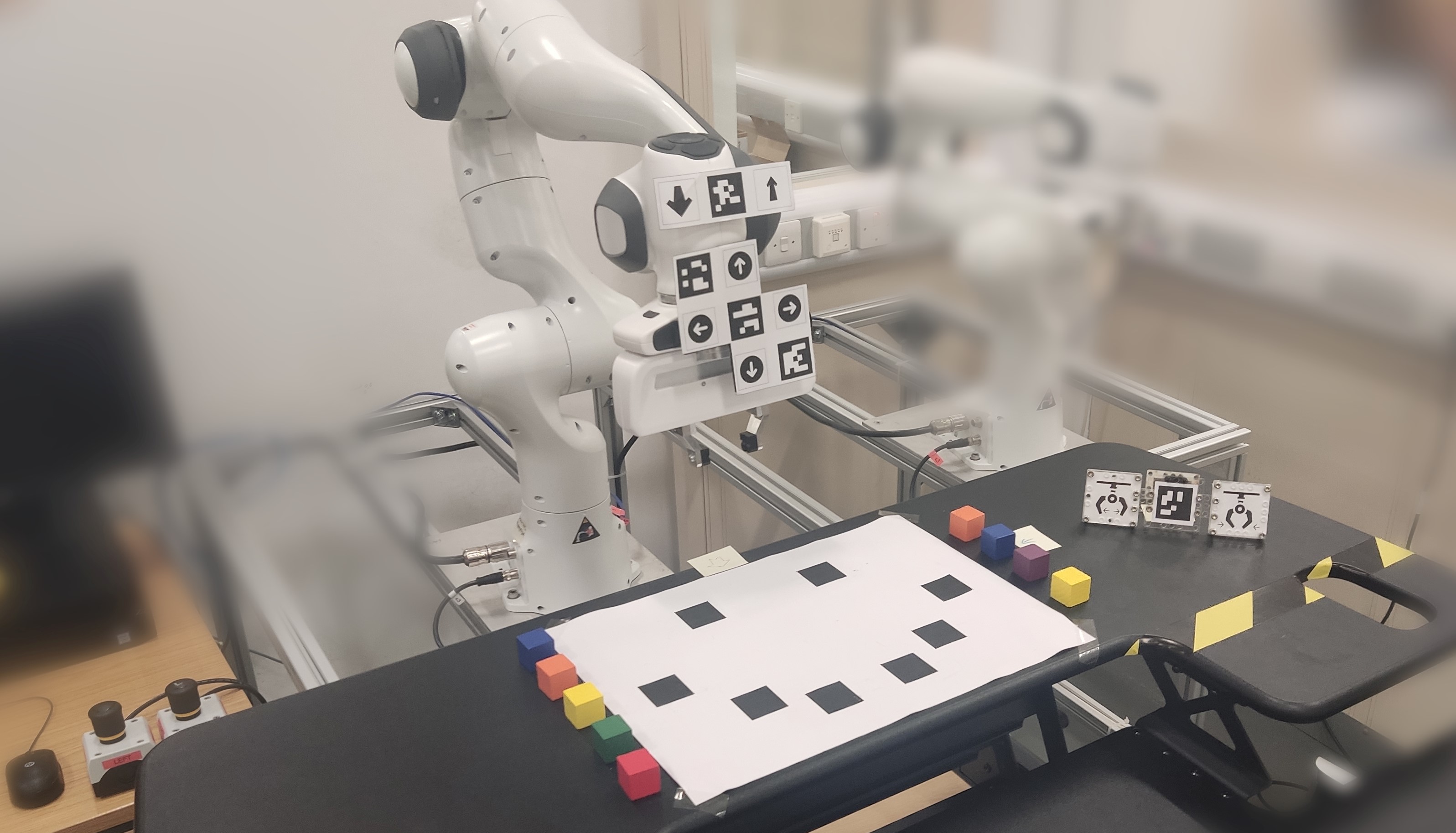}
    \hfill
    \caption{User testing setup overview.}
\label{fig:RobotSetup}
\end{figure}

The setup shown in Figure \ref{fig:RobotSetup} was prepared. 
The user sits down in front of a height-adjustable table, with the workspace area marked. 
The Panda robot arm is set opposite from the seated participant, roughly 80 cm from them. 
When the system is operational, the user's head is at roughly 50 cm from the end effector.
The table contains the stencil from the YCB Block Pick and Place benchmarking protocol. 
The virtual barrier end zones of the robot have been marked for user safety with yellow tape. 
The buttons controlling the closing or opening of the end effector are placed on the table next to the cubes.


\subsection{Experimental Procedure}

Participants were informed of the rules and scoring system.
Emphasis was placed on maintaining a high score (demonstrating high dexterity) as the goal of the test, but that time would also be recorded. 
The users were shown a brief real-time view of the glasses' external camera, received an explanation of the working principle of the system, and were informed that the robot can only be controlled if the interactive elements are in the view range of the camera, and the optimal area for interaction -roughly aligned at eye level.
Participants were informed that they were welcome to stop the experiment at any point if they felt uncomfortable or nauseous.

Before starting, it was checked if the participants were successfully recognized by the gaze-tracking glasses. 
After arriving the participants were instructed to remove eye wear except for contact lenses. 
The Tobii Pro Glasses 2 correction lenses were offered to participants for minor vision correction during the experiment.  
The participants had to undergo the proprietary calibration for the Tobii Pro Glasses 2 as described in its user manual. 
While wearing the glasses, the user is instructed to look at a card with concentric circle markers, which is gently moved during calibration. 
If the procedure could not be accomplished the test was concluded. 
Following this, the experimental protocol is divided into 2 phases: familiarization with the control system and the experimental session.

If the calibration was successful, a brief familiarization routine was conducted.
The user was instructed to look at each of the buttons to move the robot in a certain direction, ensuring the robot was working properly.
If the user was able to properly control the robot, a cube was placed in the centre of the template, and they were instructed to pick it up, raise it and drop it, concluding the familiarization process.

It is known that the Tobii Pro 2 Glasses present drift error over time, resulting in less precise gaze position identification. 
If the user reported that the robot was irresponsive during testing, they were invited to pause the test and re-run the calibration procedure.
During this, the timer was stopped and after recalibration the exercise continued.  
The robot presented some issues when approaching the cubes at the edge of the stencil closest to its base. 
If the user tried to push the robot further back it would get close to singularity positions, resulting in slowdown. 
If the user moved the robot to more extreme positions, this would cause the end effector to drift, twisting it and leaving it no longer comfortably perpendicular to the user's field of view. 
If this occurred, the test was briefly paused and the end-effector quickly re-aligned. 
If a user triggers the robot emergency stop by forcing the end effector towards the table, the timer is paused, the robot reset, and the test resumed.

\subsection{Participants}


Students and staff from the University of Bristol and the University of the West England were recruited and invited to participate.
Inclusion criteria were having working stereoscopic vision (i.e. two healthy eyes) and no severe visual impairments, here defined as being able to see the button symbols on the robot. 
The following supplemental materials were provided to the participants: A consent form to be signed, a participant information sheet, a privacy notice, and a notice of ethical approval.
Participation was voluntary, and no incentive was given.
The project was reviewed and approved by the University of the West of England University Research Ethics Committee (REF No: FET-2122-77). 

\begin{table}[tb]
    \caption{Self-reported User demographics.
    Note that the group of glasses users is not the inverse of participants that reported perfect vision. 
    i.e. some users with sight issues do not wear glasses, and vice versa.
    }
    \label{Tab:UserDemo}
    \centering
\begin{tabular}{llll}Total            & Category                   & 21 & 100.0\% \\ \hline
Gender           & Male                       & 14 & 66.7\%  \\
                 & Female                     & 6  & 28.6\%  \\
                 & Non-binary                 & 1  & 4.8\%   \\ \hline
Age              & 20-29                      & 14 & 66.7\%  \\
                 & 30-39                      & 3  & 14.3\%  \\
                 & 40-49                      & 4  & 19.0\%  \\ \hline
Ethnicity        & White British              & 12 & 57.1\%  \\
                 & White other                & 2  & 9.5\%   \\
                 & Mixed                      & 2  & 9.5\%   \\
                 & Latin American             & 1  & 4.8\%   \\
                 & Indian                     & 1  & 4.8\%   \\
                 & Persian                    & 1  & 4.8\%   \\
                 & Arab                       & 1  & 4.8\%   \\
                 & Black                      & 1  & 4.8\%   \\ \hline
Perfect Vision   & 20/20 Vision               & 11 & 52.4\%  \\
                 & Hyperopia (Farsightedness) & 2  & 9.5\%   \\
                 & Myopia (Nearsightedness)   & 1  & 4.8\%   \\
                 & Myopia and Hyperopia       & 2  & 9.5\%   \\
                 & Astigmatism and myopia     & 2  & 9.5\%   \\
                 & Astigmatism                & 2  & 9.5\%   \\
                 & Lazy eye                   & 1  & 4.8\%   \\ \hline
Corrected Vision & Glasses                    & 9  & 42.9\%  \\
                 & Contact lenses             & 1  & 4.8\%   \\
                 & No glasses                 & 11 & 52.4\% 
\end{tabular}
\end{table}

Twenty-three participants were recruited, out of which twenty-one were able to participate and complete the experiment.
From the recruited group, two participants were unable to participate: one was not properly recognized by the glasses' proprietary software, and one was rejected by the glasses' calibration process. 
Participants answered demographic questions covering Gender, Age, Ethnicity and self-reported quality of vision, including whether the participant uses glasses, with responses included in Table \ref{Tab:UserDemo}.

\subsection{Performance Metrics \& Questionnaire}


To measure usability, workload and limitations of the system, a questionnaire was administered after the test comprised of the System Usability Scale (SUS) questionnaire \cite{BrookeSUS1995}, five additional questions following the same Likert-scale format targeting specific system features, the NASA Task Load Index (NASA-TLX) assessment with weights \cite{hartDevelopmentNASATLXTask1988}, and 6 open questions regarding improvements and limitations. 
The SUS and NASA-TLX are commonly used scales that provide a convenient way to compare against other assistive devices, gaze-controlled or not.
The NASA-TLX has also been used to evaluate complex gaze-operated tasks, particularly on surgical applications \cite{fujiiGazeContingentCartesian2013,fujiiGazeGestureBased2018}.
Together with the YCB final score, these act as the final outcome variables.
The SUS questionnaire covers the main factors of usability and is complemented with the adjective scale \cite{bangorDeterminingWhatIndividual2009}, a seven-point scale which associates adjectives to average scores: 
Worst Imaginable ($12.5$), 
Awful ($20.3$), 
Poor ($35.7$), 
Ok ($50.9$), 
Good ($71.4$), 
Excellent ($85.5$), 
Best imaginable ($90.9$).  
The SUS was supplemented with 5 additional questions following the same format targeting specific elements of the system: the comfort of the glasses (11), responsivity of the system (12), interface intuitiveness (13), feelings of nausea/discomfort (14) and fatigue (15).
The NASA Task Load Index measures workload across six scales (Mental Demand, Physical Demand, Temporal Demand, Performance, Effort, Frustration). 
Participants perform pairwise comparisons of all measured scales, and the number of times each is chosen is the associated weight.
Individual scale values are measured from 0 to 20, and the weighted score corresponds to the weighted average using the comparison values as weights, from 0 to 100.
The Raw Task Load Index (RTLX) \cite{hartNasataskLoadIndex2006} is also included as it is reported by other publications featuring robot control strategies, comprised of the average score without the weights.  
The open questions target the elements mentioned in the Block Pick and Place Benchmark, asking for detailed feedback on well-performing areas, limitations, changes in the interface and the robot, and recommendations for features and applications. 
The final question asks whether the participant has familiarity with similar eye-controlled systems, to control and establish previous knowledge which could affect the learning process.

Values have been reported as $MEAN, (\pm STD)$.
Precision has been limited to two decimal positions.
{The Shapiro\-Wilk test was used to determine if scores were normally distributed, the Mann-Whitney U test to determine if two small and non-normal distributions were different, and the Spearman's rank correlation test was used to evaluate the monotonicity of relationships.}  
A p-value of $p<0.05$ was considered significant.



\section{Results}
\label{s:Results}



\subsection{YCB Block Pick and Place Protocol}
\label{ss:YCBScoring}

\begin{figure}[th]
    \centering
    \subfloat[Score Distribution for YCB Block Pick and Place Protocol \label{fig:YCBScoreHist}]{\includegraphics[width=0.23\textwidth]{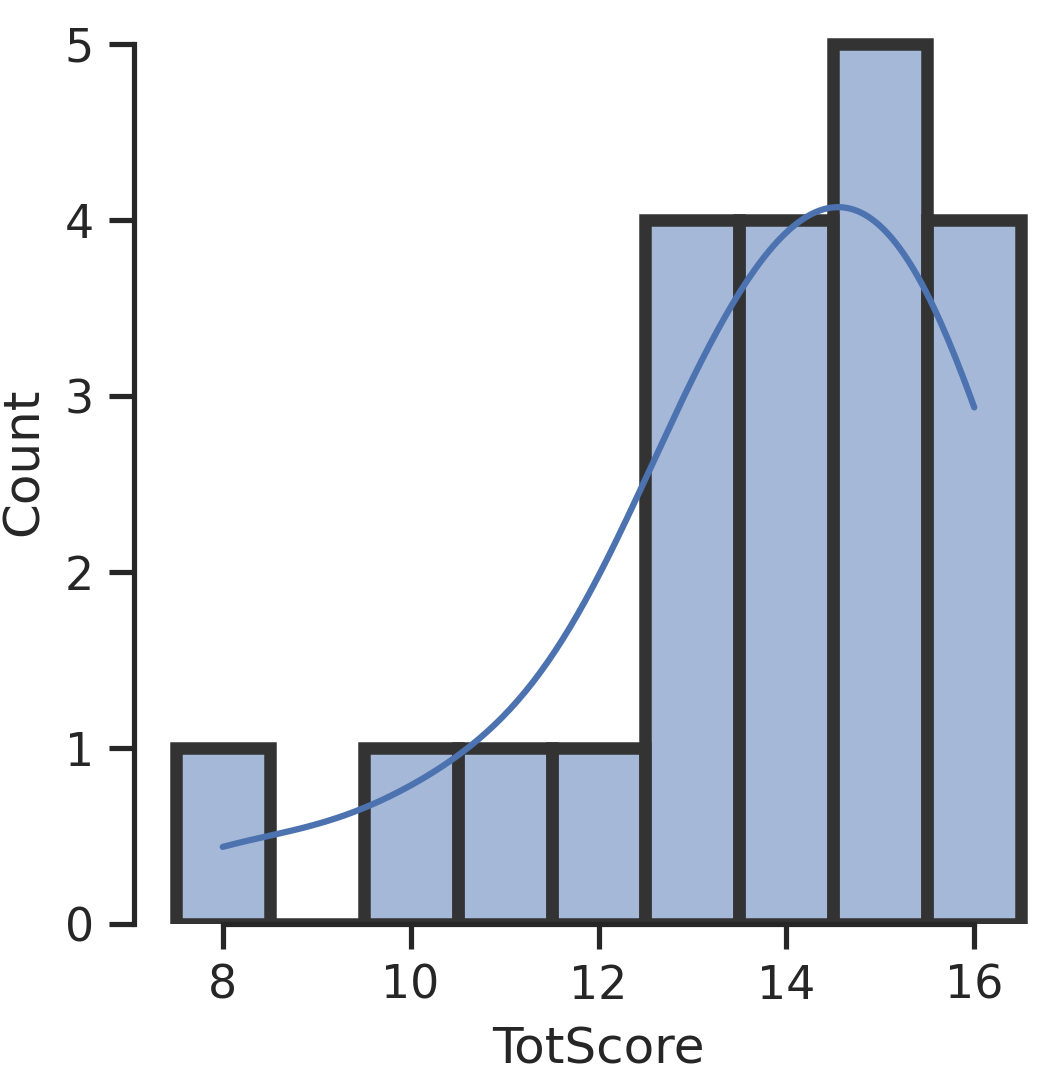}}
    \hfill
    \subfloat[Score evolution for each user with time. Slope corresponds to points per time measure. \label{fig:YCBScoreVTime}]{\includegraphics[width=0.235\textwidth]{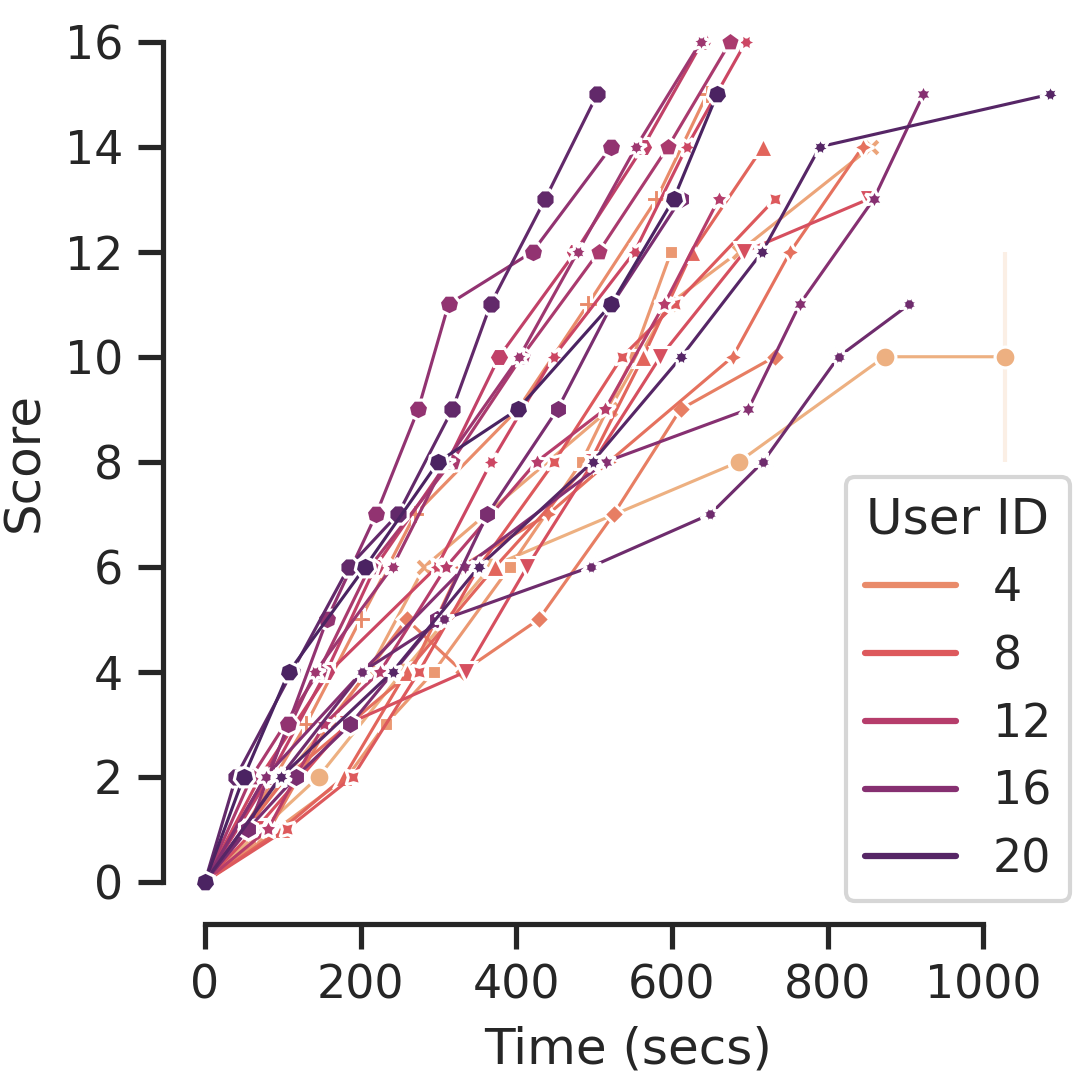}}
    \hfill
    \subfloat[Duration of each of the eight pick and drop actions. 
    The whisker reflects the extrema and each box tick the 25th percentile, median, and 75th percentile.  \label{fig:YCBTimePerAction}]{\includegraphics[width=0.490\textwidth]{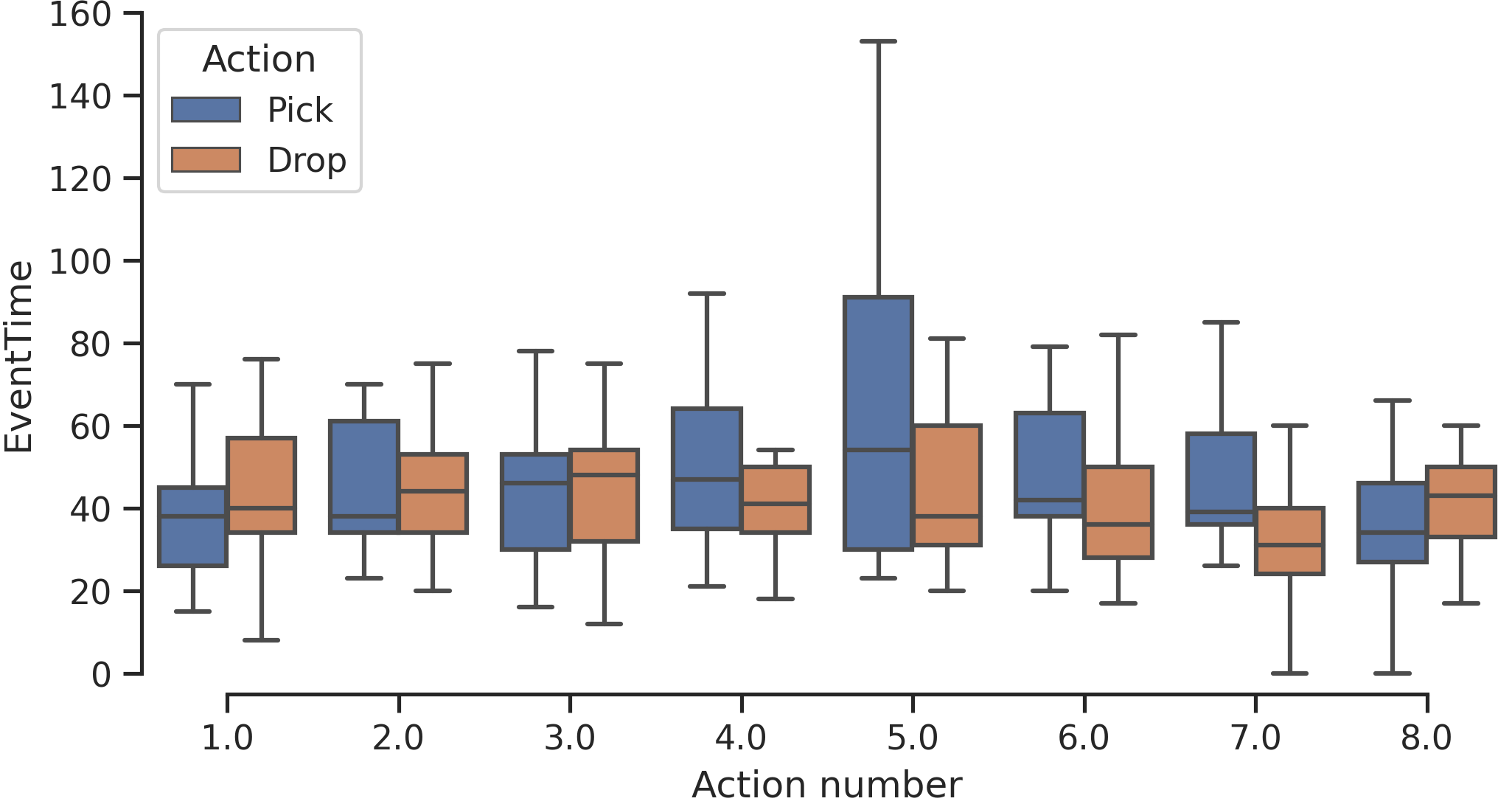}}
    \caption{YCB Results, broken down by pick-up and drop-off time.}
    \label{fig:ScoreFigs}
\end{figure}

Score information is summarised in Figure \ref{fig:ScoreFigs}.
The System Usability Score of $75.36$ places the system between "Good" ($71.4$) and "Excellent" ($85.5$) in the adjective rating scale \cite{bangorDeterminingWhatIndividual2009}. 
The mean score of $13.71 (\pm 2.10)$ points, with an execution time for the whole protocol of $739.10 (\pm 156.52)$ seconds, or $12$mins $19.10$sec $(\pm2$min $36.52$sec$)$.
Complete score distribution can be found in Figure \ref{fig:YCBScoreHist}, and scores broken down by outcome can be found in the YCB form in Table \ref{Table:YCBForm}. 
Applying a Shapiro–Wilk test points towards the scores not being normally distributed ($\text{p-value} = 0.015 < 0.05$).



\newcolumntype{C}[1]{>{\centering\let\newline\\\arraybackslash\hspace{0pt}}m{#1}}

\begin{table}[bt]
\label{Table:YCBForm}
\caption{Complete benchmark results for YCB Block Pick and Place Protocol \cite{BenchmarkingManipulationResearchUsingYCBObjectandModelSet_Calli_2015}, sorted by final score. 
Table shows the number of cubes on each target.
A deeper shade represents a higher number in each column.
} 
\centering
\begin{tabular}{C{2em}C{4em}C{4em}C{4em}C{4em}C{3em}} 
\hline
                       & \multicolumn{4}{c}{Blocks on position}                                                                        &                            \\ \cline{2-5}
\multirow{-2}{*}{User} & Entirely on target        & Partially on   target     & Partially off   template  & Entirely off   template    & \multirow{-2}{*}{Score}    \\ \hline
1                      & \cellcolor[HTML]{63BE7B}8 & \cellcolor[HTML]{FCFCFF}0 & 0                         & 0                         & \cellcolor[HTML]{5A8AC6}16 \\
2                      & \cellcolor[HTML]{63BE7B}8 & \cellcolor[HTML]{FCFCFF}0 & 0                         & 0                         & \cellcolor[HTML]{5A8AC6}16 \\
3                      & \cellcolor[HTML]{63BE7B}8 & \cellcolor[HTML]{FCFCFF}0 & 0                         & 0                         & \cellcolor[HTML]{5A8AC6}16 \\
4                      & \cellcolor[HTML]{63BE7B}8 & \cellcolor[HTML]{FCFCFF}0 & 0                         & 0                         & \cellcolor[HTML]{5A8AC6}16 \\
5                      & \cellcolor[HTML]{82CB96}7 & \cellcolor[HTML]{DEF0E5}1 & 0                         & 0                         & \cellcolor[HTML]{ABC3E3}15 \\
6                      & \cellcolor[HTML]{82CB96}7 & \cellcolor[HTML]{DEF0E5}1 & 0                         & 0                         & \cellcolor[HTML]{ABC3E3}15 \\
7                      & \cellcolor[HTML]{82CB96}7 & \cellcolor[HTML]{DEF0E5}1 & 0                         & 0                         & \cellcolor[HTML]{ABC3E3}15 \\
8                      & \cellcolor[HTML]{82CB96}7 & \cellcolor[HTML]{DEF0E5}1 & 0                         & 0                         & \cellcolor[HTML]{ABC3E3}15 \\
9                      & \cellcolor[HTML]{82CB96}7 & \cellcolor[HTML]{DEF0E5}1 & 0                         & 0                         & \cellcolor[HTML]{ABC3E3}15 \\
10                     & \cellcolor[HTML]{A1D7B0}6 & \cellcolor[HTML]{BFE4CB}2 & 0                         & 0                         & \cellcolor[HTML]{FCFCFF}14 \\
11                     & \cellcolor[HTML]{A1D7B0}6 & \cellcolor[HTML]{BFE4CB}2 & 0                         & 0                         & \cellcolor[HTML]{FCFCFF}14 \\
12                     & \cellcolor[HTML]{A1D7B0}6 & \cellcolor[HTML]{BFE4CB}2 & 0                         & 0                         & \cellcolor[HTML]{FCFCFF}14 \\
13                     & \cellcolor[HTML]{A1D7B0}6 & \cellcolor[HTML]{BFE4CB}2 & 0                         & 0                         & \cellcolor[HTML]{FCFCFF}14 \\
14                     & \cellcolor[HTML]{BFE4CB}5 & \cellcolor[HTML]{A1D7B0}3 & 0                         & 0                         & \cellcolor[HTML]{FBE3E6}13 \\
15                     & \cellcolor[HTML]{BFE4CB}5 & \cellcolor[HTML]{A1D7B0}3 & 0                         & 0                         & \cellcolor[HTML]{FBE3E6}13 \\
16                     & \cellcolor[HTML]{BFE4CB}5 & \cellcolor[HTML]{A1D7B0}3 & 0                         & 0                         & \cellcolor[HTML]{FBE3E6}13 \\
17                     & \cellcolor[HTML]{BFE4CB}5 & \cellcolor[HTML]{A1D7B0}3 & 0                         & 0                         & \cellcolor[HTML]{FBE3E6}13 \\
18                     & \cellcolor[HTML]{DEF0E5}4 & \cellcolor[HTML]{82CB96}4 & 0                         & 0                         & \cellcolor[HTML]{FACBCD}12 \\
19                     & \cellcolor[HTML]{DEF0E5}4 & \cellcolor[HTML]{A1D7B0}3 & \cellcolor[HTML]{FCE4D6}1 & 0                         & \cellcolor[HTML]{F99A9C}10 \\
20                     & \cellcolor[HTML]{FCFCFF}3 & \cellcolor[HTML]{63BE7B}5 & 0                         & 0                         & \cellcolor[HTML]{FAB2B5}11 \\
21                     & \cellcolor[HTML]{A1D7B0}6 & \cellcolor[HTML]{FCFCFF}0 & 0                         & \cellcolor[HTML]{FCE4D6}2 & \cellcolor[HTML]{F8696B}8  \\ \hline
\textbf{Avg}           & \textbf{6.10}             & \textbf{1.76}             & \textbf{0.05}             & \textbf{0.10}             & \textbf{13.71}            
\end{tabular}
\end{table}

One of the participants stopped the experiment midway due to eye strain, with two blocks not positioned.
These remaining blocks were counted as misplaced, subtracting two points each from the participant's final score, and their results were not included in the timing calculations. 
The robot emergency stop was triggered by one user. 

The actions to achieve the task were classified and split on pick-up and drop-off for each individual block, shown in Figure \ref{fig:YCBTimePerAction}. 
On average from start to finish, pick-up took $44.55 (\pm 23.24)$ secs and dropping-off $48.55 (\pm 28.56)$ secs. 
As seen in the figure, the fifth pick action was slower, as participants usually had to move the robot arm from one end of the template to the opposite to resume pick-up.  
Cubes farther away from the users were usually the last ones reached during the experiment. 
Figure \ref{fig:YCBScoreVTime} shows the score for each participant changing with time, with  participants taking an average of $69.61$ $(\pm1.78)$ seconds to win a point.









To address possible individual differences or discrimination present in the system, the YCB scores were compared among groups based on demographic groups. 
Results have been included in Table \ref{Tab:YCBScoresDemo}. 
Fourteen participants identified as male, six as female, and one as non-binary. 
Due to being a single sample, the non-binary participant was excluded from the gender split.
No significant difference was found between female $14.16 (\pm 1.47)$ and male $13.35 (\pm 2.31)$ 
($\text{p-value} = 0.321$).
To explore the effect of age, participants were divided by the median age (28 years) into the groups "equal or older than 28" and "younger than 28", but no significant difference was found ($\text{p-value} = 0.86$) for the final respective scores of $13.54 (\pm 2.42)$ and $13.90  (\pm 1.79)$. 
Regularly wearing glasses $14.0 (\pm 2.0)$ did not seem to affect scores significantly when compared to non-wearers $13.45 (\pm 2.25)$ ($\text{p-value} = 0.605 $). 
While not statistically significant ($\text{p-value} = 0.223 $), users that reported an eye condition or defect - ranging from  minor prescription lenses to astigmatism and lazy eye - showed the largest discrepancy, with scores of $14.45 (\pm 1.13) $ for self-reported perfect sight against  $ 12.90 (\pm 2.64) $ for users with self-reported issues. 
To control for digital discrimination/bias, scores were compared separated by race, with no significant difference observed ($\text{p-value} = 0.97$) between ethnic white users $13.77 (\pm 2.12)$ and non-white users $13.63 (\pm 2.20$).



\begin{table}[tb]
    \centering
    \caption{YCB test scores after controlling for demographic factors. 
    p-value shown for the Mann-Whitney U test.
    }
    \label{Tab:YCBScoresDemo}
\begin{tabular}{llcc}
Category          & Comparison                        & YCB Score                    & p-value     \\ \hline
Gender         & Male                          & $13.35 (\pm 2.31)$        & 0.321       \\
               & Female                        & \multicolumn{2}{l}{$14.16 (\pm 1.47)$}  \\ \hline
Age            & Age\textgreater{}=28          & $13.54 (\pm 2.42)$        & 0.86        \\
               & Age\textless{}28              & \multicolumn{2}{l}{$13.90  (\pm 1.79)$} \\ \hline
Glass Wearers  & Wearer                        & $14.0 (\pm 2.0)$          & 0.605       \\
               & Non-wearer                    & \multicolumn{2}{l}{$13.45 (\pm  2.25)$} \\ \hline
Vision & Self-reported issues                        & $ 12.90 (\pm 2.64) $      & 0.223       \\
               & 20/20 Vision                    & \multicolumn{2}{l}{$14.45 (\pm 1.13) $} \\ \hline
Ethnicity      & White British & $13.77 (\pm 2.12)$        & 0.97        \\
               & Other groups                  & \multicolumn{2}{l}{$13.63 (\pm 2.20$)} 
\end{tabular}

\end{table}

\subsection{System Usability Scale}

\begin{figure}[thb]
\centering
\includegraphics[width=0.49\textwidth]{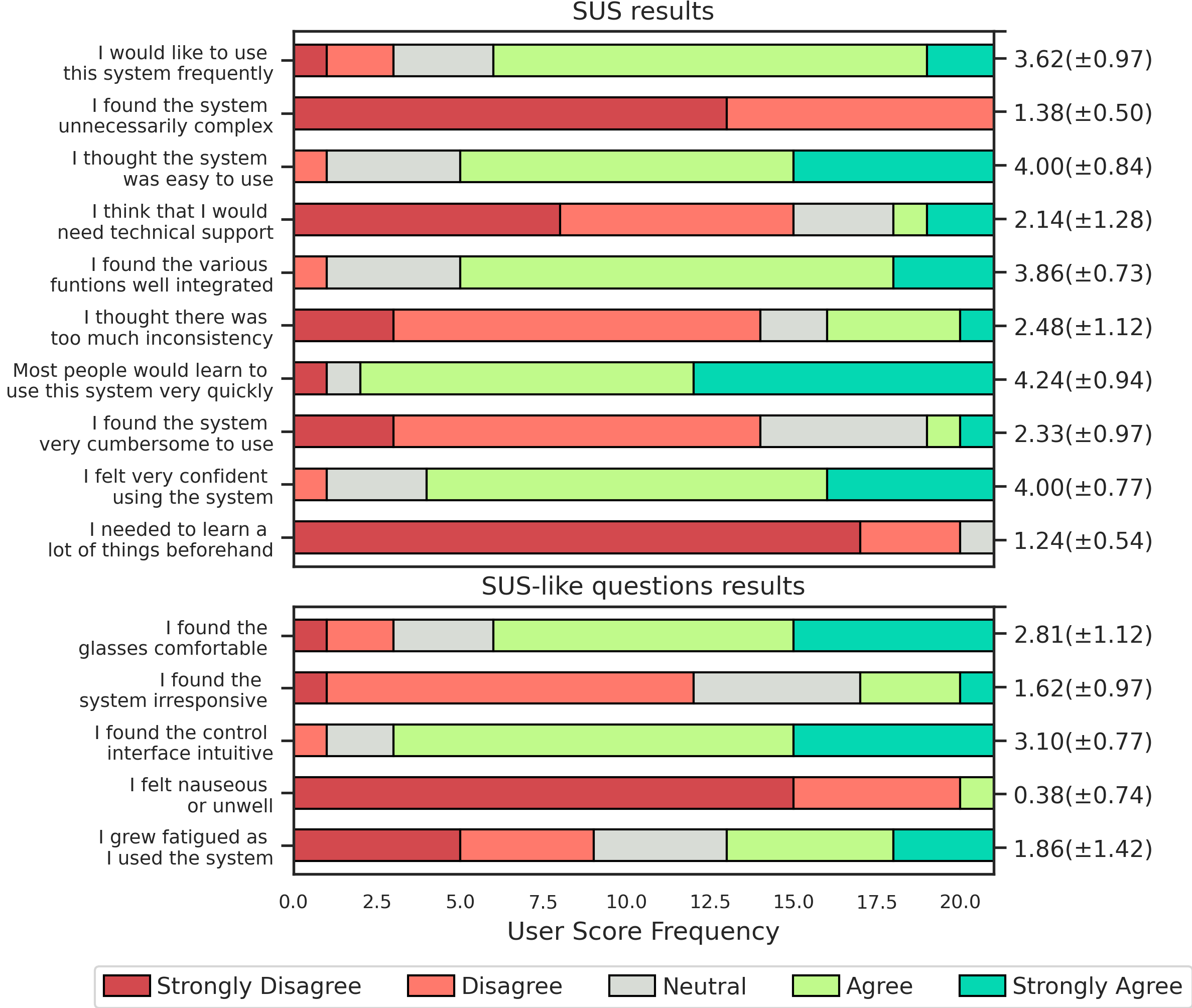}
\caption{Frequency of results for the System Usability Scale (SUS Q1 to Q10 in order) and our SUS-like questions.
Averages and std-dev are shown on the right side. 
For the SUS, odd-numbered questions refer to positive sentiment, even-numbered ones are negative.}
\label{fig:SUSResults}
\end{figure}

Figure \ref{fig:SUSResults} shows the distribution of the results for the System Usability Scale test (mean score of $75.36 (\pm 13.26)$) and the additional SUS-like questions, 

Most participants found the system to be highly intuitive to use, with two neutral and one disagreeing.
This matches the experimental observations, as participants learned how to use the system very quickly, requiring minimal instruction.
Several attempted to grab the practice block as soon as they received control of the robot before the familiarization phase had started.  
Overall responses seem positive, with more than 80\% of participants agreeing or strongly agreeing that the system had low complexity (Q2), was quick to learn (Q7), that they felt confident while using it (Q9), and that they needed to learn few things beforehand (Q10).

Responses to the additional SUS-like questions were found to be mixed. 
The interface was reported as intuitive, and the majority of users found the glasses comfortable to wear during the test. 
Feelings of nausea or discomfort were low, only one participant reported feeling nauseous after the experiment (Q14).
However, fatigue (Q15) was much more noticeable, with 8 participants reporting some degree of it and 3 feeling very fatigued, matching the user comments regarding eye strain. 
Fatigue showed the largest standard deviation of all questions $ 1.86 (\pm 1.424279)$, with answers uniformly distributed across the spectrum.






\subsection{NASA Task Load Index}

\begin{figure}[tb]
     \centering
        \subfloat[]{\includegraphics[width=0.48\textwidth]{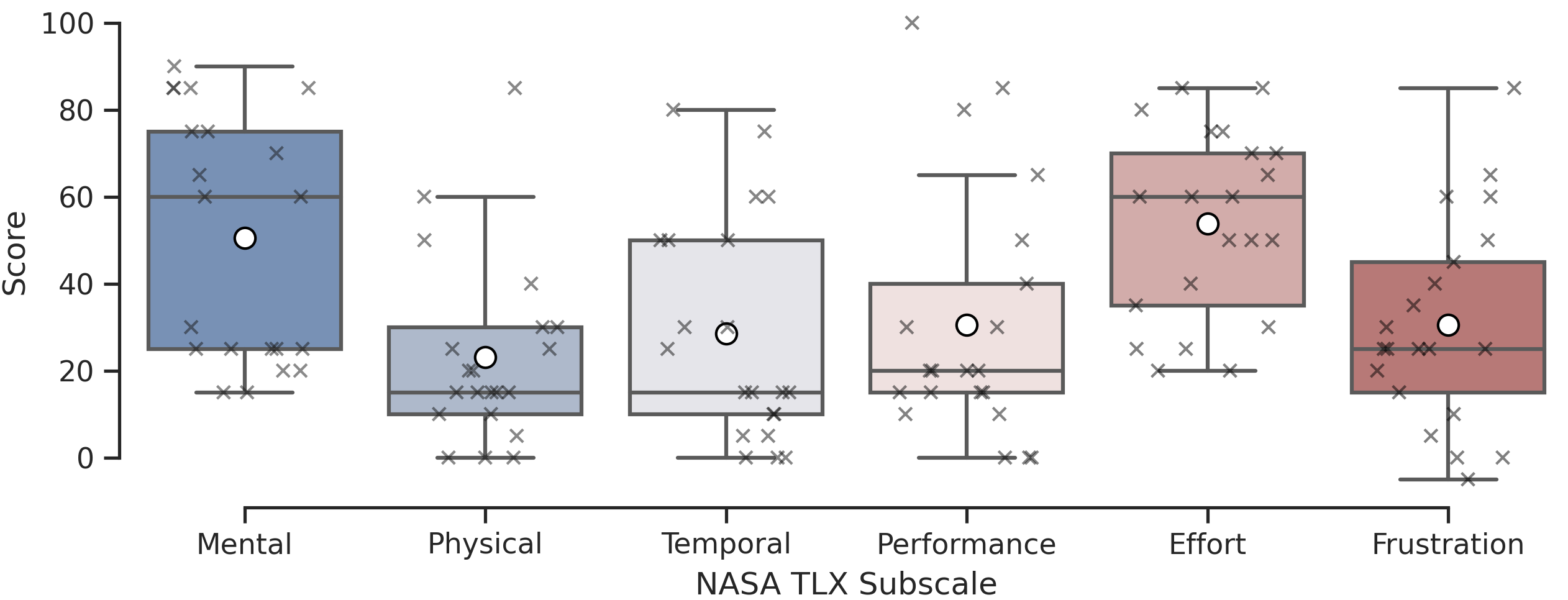}}
     \hfill
        \subfloat[]{\includegraphics[width=0.48\textwidth]{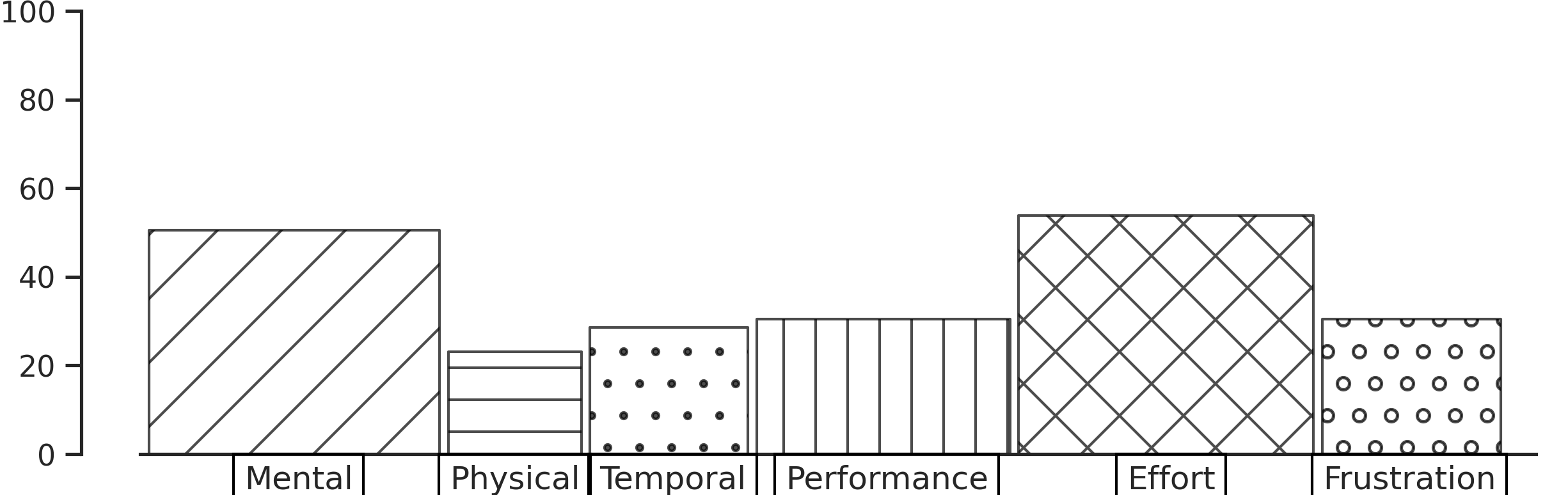}}
    \caption{NASA-TLX results for each category showing (a) Distribution plots for unweighted scores, where box shows the interquartile range and whiskers show the rest of the distribution excluding outliers \cite{Waskom2021}. (b) Mean weighted scores. The width corresponds to the weight reported by the users.}
    \label{fig:NASATLX}
\end{figure}

The average score for NASA-TLX total workload was $44.76 (\pm 18.77)$, with an average unweighted/raw score of $36.15 (\pm 12.72$).
NASA TLX subjective workload score distribution for each category is shown in Figure \ref{fig:NASATLX}.
Mental demand (50.48) and effort (53.81) were reported as the highest contributors to workload. 
Physical (23.10) and temporal (28.57) demand was reported to be the smallest.

\subsection{Qualitative Assessment: Comments and Open Questions}

Users were observed to move their heads during the experiment to keep the robot end-effector in view perpendicular to the glasses. 
Some users were observed to exaggerate their movements to prevent triggering actions accidentally, drawing a wide arc with their heads, particularly noticeable when trying to open or close the grasper. 

Brief conversations with the users were had after using the device, where users described the system in favourable terms, as intuitive, self-explanatory, very responsive, and reported feeling confident with the system.
Here answers to the open questions are summarised and grouped by question, and then theme when considered suitable.
Illustrative comments were included for each question.


\subsubsection{Q1. If you had to use an assistive robot for physical tasks, would you use this system?}


When asked whether they would like to use this system for assistive tasks if they had a severe disability, 12 responded affirmatively without conditions, and 7 pointed out that the system needed improvements in responsiveness and habituation before being used properly. 
Mental fatigue and the need for practice were identified by a participant as a significant obstacle. 
Issues arising due to the inverse kinematics were mentioned by one participant.
Two participants proposed potential applications of D-GUIs aside from robot control, such as turning on the lights or a TV remote.


\begin{quote}
    \textit{"I found it intuitive and relatively easy to use."} \\
    \textit{"It was simple to use and required little effort on my part."}
\end{quote}

\subsubsection{Q2. What worked well about the interface? What should be changed?}






Participants praised the system interface (x9 out of 21), calling it responsive (x4), mentioning that it worked well (x3) and feeling confident that they would not make mistakes while using it (x2). 
Participants described the system as innovative, easy to understand, self-explanatory, simple, intuitive, and in need of no changes (x1 each).
Participants were critical of the gripper controls being located away from the rest of the robot (x4), mentioning accidental inputs when switching (x2).
Confusion over the interface symbols for the depth controls (x2), 
accidental inputs such as the down arrow being too close to the end effector (x2), and needing to move their eyes outside the scene while thinking about their next action (x1) were the most common criticisms. 
Users reported wanting to customize the interface in some way, by switching the positions of the close-far and up-down buttons,  
disconnecting the open/close buttons  (x1 each).
One participant tried to move the interface elements mid-test. 
The need for facial alignment or head movement was mentioned as a barrier to serving a paralysed population (x2).

\begin{quote}
    \textit{"A way of pausing the motion would be good to make it easier to check the cube’s current location."} 
\end{quote}
\begin{quote}
    \textit{"The interface requires head and body movement in order to be able to receive the commands.  This should be changed for the final application."}
\end{quote}

\subsubsection{Q3. What worked well about the robot? What should be changed?}






As for the robot response, users felt that it was safe to use (x3), robust against impacts (x2), and reported liking the simple gripper (x2).
Some alluded to slow movement/tilt caused by movement close to the singularity position (x2) and the drift that could result from long operation (x7).
Precision issues when aligning the cubes requiring adjusting the position repeatedly before dropping (x4) were a common criticism. 
Thoughts on speed were mixed, some users commented favourably on the speed of the robot (x2), some requested it to be increased (x4) and some wanted direct means of speed control (x2) 
to switch between a fast 
and slow mode. 
A user desired the robot to keep accelerating continuously as the user continuously commanded. 
Participants commented on the "smoothness" of the robot movement, describing it as "very smooth" (x4), and "jerky" (x2), describing perceived choppy behaviour. 
One user commented that the axis of movement of the robot was not perfectly aligned with the stencil, leading to mental strain. 

\begin{quote}
    \textit{"The robot moved and behaved as expected. It didn’t crash when there was contact, which made it robust to error. I think this is a good feature as I didn’t have to reset it when I did an incorrect action."} 

    \textit{"[...] It could move faster or with variable speed, slower when precision is required, faster in transit."}
\end{quote}

\subsubsection{Q4. What other functions would you like to see on the system?}


When asked about additional features, users wanted control of the movement speed (x5), control of rotation to be implemented (x2), 
Additional feedback on the position of the gripper while moving (x2) was requested, and proposed via haptic systems (x1).
A greater degree of control over the end effector claw was also mentioned (x1). 
Visual feedback from the buttons was mentioned as well, equipping them with simple lights to indicate when they are being interacted with (x1). 
Proposed features included a pause button to disable movement and allow studying the scene without risk of accidental inputs (x2), 
a dedicated option or button to reset the arm back to a base position (x2), 
switching the buttons into toggles making the end-effector remain in motion until stopped by the user (x1), 
and using external devices to trigger the buttons after looking (x1). 
More advanced proposals included the ability to program individual positions or actions to teach the robot (x1), and the ability to autonomously pick items. 
Options to automatically align the interface with the user's face were also mentioned.

\begin{quote}
    \textit{"Ability to program the robot to repeat movements (like saving a module for an action)."}
\end{quote}

\begin{quote}
    \textit{"Feedback of the position of the gripper. [As I] Can't see it directly"} 
\end{quote}


\subsubsection{Q5. What specific daily life task would you use this for?}


Some \textit{Activities of Daily Living} were suggested as possible with the system, including picking and rearranging everyday items (x7), making or consuming food (x4), dressing (x2), washing (x2), or applying makeup (x1). 
Other tasks like opening cupboards or doors (x2), putting shopping away (x1) or washing (x1) were also suggested. 
A participant felt confident the robot was capable of performing everyday tasks already. 

\subsubsection{Q6. Do you know other similar systems?}

Participants reported not knowing of similar systems. 
One reported having worked with dry-EEG for visually evoked brain potentials, but not in an assistive context.

\section{Discussion}
\label{s:Discussion}


In this article, we demonstrate a novel screen-less gaze-controlled interface system, and its application to freely control a 3DOF robotic arm to accomplish a complex pick-and-place task requiring high precision.  
The interface can be placed directly onto the robot to allow intuitive and direct interaction and eliminate the need for context-switching between external screens and the robot.
It also allows to use of more and larger interactive elements, reducing the need for precise gaze systems. 
Unlike on-screen gaze systems, D-GUI also allows for individualization, further reinforced by the user who attempted to modify the interface mid-test. 
During the test, users were able to control the system immediately, requiring minimal explanations and habituation time.
Results show that users can use the multiple interface elements to perform continuous and discrete actions with minimal practice and with a small cognitive load.
Furthermore, the system is tested on a diverse set of users from multiple backgrounds, ages and vision quality conditions. 
To our knowledge, this is also the first system that enables continuous end-effector control in 3-D space purely via gaze, with no gestures required.

\subsection{YCB Block Pick and Place Protocol}


Participants obtained high scores in the \textit{Yale-CMU-Berkeley (YCB) Block Pick and Place Protocol} ($13.71$ out of $16.00$), showcasing how the interface and system can be used to execute a task requiring precision and dexterity. 
Observing the change of scores with time (Figure \ref{fig:YCBScoreVTime}), after normalizing by the maximum time for each user no noticeable change in slope was found for the score evolution (Linear regression $R^2= 0.909, MSE= 1.96$, quadratics regression $R^2= 0.910, MSE= 1.95$), suggesting no changing rate in user response.
Thus, participants placed the cubes at the same rate throughout the experiment. 
As some participants reported some degree of fatigue and were observed to place the most readily available cubes first and to leave the most distant cubes to place to the end, this could suggest speeding up, and thus some degree of learning while operating the system, but dedicated testing is needed to ascertain this.

Excluding the penalty due to incomplete exercise, only one mistake occurred during testing for all pick-drop actions, with a participant accidentally triggering a block release partially outside the template area.
While not conclusive, it points to the discrete approach for the gripper as robust, particularly against accidental drops. 
False positives for continuous actions are more difficult to quantify. 
On the other hand, while the continuous control approach used maximises usability with brief dwell times, it may result in increased accidental actions, evidenced by one participant triggering an emergency stop by applying excessive force against the table. 
Exploring the ideal values for the underlying control variables of the continuous mode ($T_c$, $a_{\text{on}}$, $a_{\text{off}}$, $v_{\text{ref}}$) is of interest, and might depend on user preference or expertise with the system requiring dedicated studies.

Occasionally, the arm moving resulted in the grasp controls stand being blocked from view, requiring it to be moved to enable grasping. 
Marker localization was recognized on average during $98.38\%$ of the frames for all users. 
While this result suggests a robust response, real illumination conditions will be different from our test conditions. 
Means to increase the robustness of detection are discussed in Section \ref{ss:FutureWorks}.





While the \textit{YCB Block Pick and Place Protocol} provides a good benchmark to test the precision of the system, tests more representative of real-world conditions should be used in the future.
For our use case, the test does not present constant difficulty because the number of available blocks goes down the number of options is reduced, thus increasing difficulty.
Moreover, the distance to different targets and workspace of the robot makes different blocks more difficult to manipulate, and it does not test manipulation with changes in height (3D manipulation). 
These factors make it difficult to observe learning effects. 
Controlling for the start and end position of each block could help control these artefacts.
For future works, it is recommended to complement standardised methods with concrete Activities of Daily Living as a bench-marking to provide a more representative use case before introducing it to users.

\begin{table*}[tb]
    \centering
    \caption{Collected results for NASA-TLX in other works for manipulation for users with tetraplegia. 
    If unspecified, it was assumed that the weighted NASA-TLX score was reported.}
    \label{Tab:NASAComparison}
\begin{tabular}{rL{8cm}C{2cm}C{2cm}}
Category                                   & Approach                                                                                            & NASA-TLX    & RTLX        \\ \hline
                                           & D-GUI for YCB protocol                                                                                               & $44.76 \pm 18.77$    & $36.15 (\pm 12.72$) \\ \hline
\multirow{5}{*}{Gaze controlled   systems} 
                                           & Active zones and gestures for 3DOF pick-up task    \cite{alsharifGazeGestureBasedHuman2016}                                 & - & $32.0 $ 
                                           \\
                                           & Gaze GUI and head movements for drink pick-up task \cite{stalljannPerformanceAnalysisHead2020}              & -                    & 24.30               \\
                                            & GUI \& Gestures. Camera control for laparoscopic surgery \cite{fujiiGazeGestureBased2018}                                         & 40.00                & -                   \\
                                           & GUI \& Gestures. Camera control for laparoscopic surgery    \cite{fujiiGazeContingentCartesian2013}                                & 41.33                & -                   \\
                                           & EOG for 1D movement   \cite{badesaPhysiologicalResponsesHybrid2019}                                 & $33.82 (\pm 14.78)$  & -                   \\
                                           \hline
\multirow{6}{*}{Other approaches}          & Joystick for standardised tasks   \cite{chungTaskOrientedPerformanceEvaluation2017}                 & $32.8 (\pm 18.9)$    & -                   \\
                                           & Keyboard for standardised tasks \cite{chungTaskOrientedPerformanceEvaluation2017}                                                                     & $39.6 (\pm 19.9)$    & -                   \\
                                           & Tongue control, 5DOF exoskeleton for 2-min drinking task \cite{mohammadiTongueControlFiveDOF2023}
 & -                    & 40.00               \\
                                           & Tongue control, 6DOF pouring task, after 2-day practice \cite{palsdottirDedicatedToolFrame2022a}
  & -                    & 23 to 35 *           \\
                                           & IMU via residual back muscle   \cite{leeExploratoryMultiSessionStudy2023}                           & 50 to 70 *             & -                   \\
                                           & EEG to open exoskeleton hand   \cite{badesaPhysiologicalResponsesHybrid2019}                        & $58.38 (\pm 16.52) $ & -        \\         
\end{tabular}
\\ \raggedleft \footnotesize{$*$ Unreported values, estimated from graphs}
\end{table*}

\subsection{SUS}

User experience was rated high in objective and subjective measures. 
Standout user results include: 
"I think the system is unnecessarily complex" where all participants disagree or strongly disagree (SUS Q2),
"Most people would learn to use the system quickly" where $19/21$ agree or strongly agree (SUS Q7), 
and "I needed to learn a lot of things beforehand" where $17/21$ strongly opposed (SUS Q10).
Comparison with other studies evaluating gaze control of assistive systems is difficult due to the heterogeneity of test conditions and lack of complete standardised benchmarks but generally places our system in favourable terms.
In the gaze-controlled assisted arm support by Shafti et al. \cite{shaftiGazebasedContextawareRobotic2019} "unpredictability" and "confidence" were reported as divisive.
These metrics match SUS Q5 and SUS Q9, which were considered favourable by our testing population.  
Our average score for \textit{ease of use} (SUS Q3) matches the value obtained by Wang et al. \cite{wangFreeView3DGazeGuided2018} in the manual mode of their gaze-controlled end effector (avg of 4.0 points for both), achieved by looking at interaction zones relative to the end-effector. 
As the YCB is more complex than the task in Wang et al. setup (placing one can in a 12x15cm box with a robot following discrete commands), results suggest that using the printed UI contributes significantly to usability, but dedicated testing would be needed.

The additional SUS-like questions show a similarly positive reception.
It is believed that the buttons being physical rather than virtual contributed to a very low sense of nausea compared to augmented reality systems, 
as there is no mismatch between virtual and real elements. 
However, response to fatigue was notably mixed, with more than a third of users reporting some degree of fatigue after the test 
commonly in gaze-controlled applications.
Some of this could be attributed to the placement of the grasping controls. 
The change in focus necessary to trigger them could be compared to the context switching/ significant gaze travel required in some gaze applications \cite{gabbardEffectsARDisplay2019}, leading to eye strain. 


\subsection{NASA-TLX}

On the NASA Task Load Index test mental demand and effort were identified as the highest contributors to workload.
It is unclear if these workload contributors, as well as the high fatigue, would diminish with regular use and practice.
The significant variation seen for Temporal demand could point towards users responding differently to prolonged use times, but more testing is needed. 
The small value for \textit{Physical demand} suggests the approach is suitable for individuals with tetraplegia or substantial disabilities.
However, it was noted that some participants moved considerably during the operation of the robot, mainly to maintain alignment with the GUI. 
Techniques to minimize the need for body movement should be implemented in future tests.
Enabling the robot to remain aligned with the user's point of view in real-time is of particular interest.

For comparison, different control systems for assistive robot systems benchmarked with the NASA-TLX system were collected, sumarised in Table \ref{Tab:NASAComparison}. 
Alsharif et al. \cite{alsharifGazeGestureBasedHuman2016} presents the most direct comparison in both system and testing strategy, demonstrating an embedded interface for controlling the robot by looking at set action zones relative to the end effector. 
The user can change between interaction modes via simple gestures to control the end-effector movement in a plane, depth, rotation or actuate the gripper.  
Participants were seated in front of the robot arm, where they had to perform two simple pick-and-drop tasks to place two cubes onto two raised platforms.
An unweighted RTLX score of $32.0$ was obtained, with the most noticeable difference being physical demand found to be 20 points higher than ours. 
This high physical demand could be attributed to gesture-based switching, requiring multiple sequences when changing manipulation mode.
This is supported by the considerable execution times in their testing, with their two pick-ups lasting approximately 12 minutes, the same as our participants in the YCB benchmarking task despite additional training time. 
The similarity in setup but the disparity in outcomes suggests that numerous or frequently used gestures can significantly slow down operation. 
Our higher overall task-load score could be attributed to a lack of initial training time, the more demanding YCB task, and faster and thus less forgiving robot parameters. 

Exploring other gaze-controlled systems from literature, Stalljann et al. \cite{stalljannPerformanceAnalysisHead2020} shows robot control via gaze-actuated GUI, in which each option guides the robot to set defined zones, but where continuous movement is then controlled via head movements. 
Eleven participants, one with tetraplegia, used the interface to pick and drink from a cup using only the discrete GUI actions.
A raw score of $24.3$ was reported for an end-effector in an item pick-up system, showing the GUI approach as less taxing.  
The more abstracted controls help user experience but also result in a more limited system depending on structured scenarios.  
Gaze interfaces have seen use in other contexts, such as for controlling laparoscopic surgery tools, shown by Fuji et al. \cite{fujiiGazeGestureBased2018,fujiiGazeContingentCartesian2013}, with weighted NASA-TLX scores of $40.00$ and $41.33$ respectively. 
Using a combination of gestures and a GUI, trained participants are able to zoom in, pan the camera \cite{fujiiGazeContingentCartesian2013} and in later work tilt using head movements \cite{fujiiGazeGestureBased2018}. 
The comparable scores do suggest that gaze interfaces can be used in highly demanding environments with precision for low workloads, but the different test conditions (use of trained specialists, use of robot as helper, robot for non-pick and place tasks) make direct comparison inappropriate.

Placing the system in a wider context outside of gaze technologies, Chung et al. \cite{chungTaskOrientedPerformanceEvaluation2017} shows users with upper extremity impairments manipulating robot arms to complete ISO 9241-9 standard tasks with joystick and keyboard controls, achieving weighted NASA-TLX scores of $32.8 (\pm 18.9)$ and $39.6 (\pm 19.9)$ respectively for each input device. 
The tasks involved pressing buttons interacting with the TO-PET interface and re-arranging items repeatedly, both representing several activities of daily living. 
Mohammadi et al. \cite{mohammadiTongueControlFiveDOF2023} reports an unweighted RTLX score $40$ of for 5DOF control of an upper limb exoskeleton for a simulated 2-minute drinking task, controlled via a tongue interface.
Similarly, Palsdottir et al. \cite{palsdottirDedicatedToolFrame2022a} uses tongue control to control a 6DOF Jaco arm from multiple reference frames in a water pouring task, scoring from 35 to 23 approximately after 3 practice sessions across multiple days. 
Lee et al. \cite{leeExploratoryMultiSessionStudy2023} opts to use residual muscle mobility, measured using 4 IMU placed on the scapulae and upper arms to control a 7DOF Jaco arm to touch 3D targets.
After five days of sessions, weighted NASA-TLX scores ranged from approximately 50 to 75 depending on control modalities.
Badesa et al. \cite{badesaPhysiologicalResponsesHybrid2019} shows naive users using EEG systems to control the opening and closing of an exoskeleton hand obtaining weighted NASA-TLX scores of $58.38 (\pm 16.52) $. 
Users were able to move the arm in 1-D via EOG signals, looking right to move the arm right, reaching scores of $33.82 (\pm 14.78)$. 
While direct comparison is difficult due to changing conditions in test, duration and users, results do suggest a similar workload of our embedded interface approach when compared to other input modalities for people with tetraplegia, particularly of the widely used traditional controllers \cite{chungTaskOrientedPerformanceEvaluation2017}. 
As mentioned by Chung et al. \cite{chungTaskOrientedPerformanceEvaluation2017} result heterogeneity also points to the need to use standardised measures for testing and reporting scores in benchmarking assistive devices.

Inter-demographic comparisons showed no statistically significant preference for any of the split groups, suggesting the system is unlikely to favour by gender, age ethnicity or eye-sight quality.
However, a larger sample of diverse participants should be used in future tests to increase confidence. 
Furthermore, future tests should increase efforts to balance the gender distribution and - as the best predictor of performance - increase the inclusion of participants with imperfect vision.
Of interest are participants with imperfect pupils, which are not considered in most gaze recognition algorithms  \cite{KarReviewAnalysisEyeGaze2017}.
The gaze-tracker used did not recognize two of the potential participants, raising the possibility of hidden digital discrimination. 
Spearman correlation test revealed a correlation between the three scores. 
A positive relationship between the YCB achieved score and perceived system usability scale was found (spearman-r $0.544$, $\text{p-value} = 0.011 < 0.05$), as well as an inverse relationship between the usability scale and the workload index (spearman-r $-0.583$, $\text{p-value}= 0.006 < 0.05$).
No significant correlation was found between the YCB score and workload index (Spearman-r $-0.381$, $\text{p-value} = 0.088$).

\subsection{Future Works and Technical Limitations}
\label{ss:FutureWorks}

\subsubsection{Interface}

The core limitation behind the Diegetic Interface lies in the dimensional constraints placed on the system to be controlled. 
As physical (non-virtual) buttons, they must be readable to the user and the eye-tracker camera simultaneously. 
The interface elements need to be large which places limits on the design of interfaces for complex systems, and compete for space with the fiducial markers. 
The system is also vulnerable to external or self-occlusion, where obstacles or user action can render the system unable to be controlled. 
This makes exploration of rotation to achieve 6DOF control challenging, and likely constraining it to face the user without hardware modifications. 
Unlike screen systems, there is a finite number of possible user actions, limiting flexibility. 
Finally, with the interface on the end effector, it is not suited to tasks that require proximity to the face of the user, such as assistive drinking. 
Such functions could be performed via automated approaches \cite{tryVisualSensorFusion2021} or multimodal systems, such as force sensors \cite{shasthaApplicationReinforcementLearning2020}.

However, as a novel approach, it also raises new possibilities not possible in traditional systems. 
Participants could easily design their own interfaces intuitively, dragging the individual elements to their liking on top of the robot, which remains the target of future qualitative studies. 
Unexplored options include embedding the markers onto the intermediate links of the robot to resolve singularities or redundant configurations or 
defining predefined actions ("shortcuts" or "macros").  
Furthermore, markers can be placed in the environment to define context-sensitive actions such as opening cabinets, interacting with appliances, operating wheelchairs or interacting with \textit{Internet-of-Things} devices. 
It is believed that the presented interface can complement existing end-point-control approaches described in Section \ref{s:RelatedWorks}, used to define and teach automatic systems grasping strategies for unknown objects or in uncertain scenarios. 
As the mismatch in position between virtual and real elements is eliminated it raises the possibility of using the diegetic elements as a baseline "screen-less virtual reality" to evaluate the effects of other head-mounted devices.  
Due to the fatigue found, more research is needed on the use of screen-less gaze-controlled devices for extended periods, and whether the lack of feedback contributes to it.


\subsubsection{Lack of Feedback}

Lack of visual feedback was a common mention by the users of the system. 
While augmented reality systems could be used to provide feedback to the users \cite{parkHandsFreeHumanRobot2021}, these carry other concerns such as known negative symptoms ("VR-sickness") and limited usability in outside everyday situations. 
Their use can also manifest an \textit{aesthetical barrier}, making the users feel alienated and least likely to use the device \cite{AssessmentofBrainMachineInterfacesfromthePerspectiveofPeoplewithParalysis_Blabe}. 
Expanding the D-GUI with interactive elements, such as a backlight, e-ink, screen or mechanical parts could allow for feedback without introducing potential sources of discomfort. 
Future systems could implement sound or light cues to communicate how much strength is being applied, separate from the interface elements.

\subsubsection{Multimodality}

To tackle the \textit{Midas touch problem} in gaze interaction: the ambiguity between intentional user interaction and accidental or exploratory eye-gazes \cite{GazebasedTeleprosthetic_Dziemian}, multimodal inputs can be used to confirm interaction. 
The small dwell times used could be a key factor leading to the good reception and performance of the system, with the slow starting velocity of the robot providing a small amount of feedback on itself. 
Using complementary signals like EMG \cite{li3DGazeBasedRoboticGrasping2017}, movement \cite{Neuronode} external switches, voice commands, sensors or gestures would ensure the intentionality of the user during interaction without the trade-off between dwell-time and usability.
Avoiding dwell times has demonstrated success in applications requiring high precision \cite{huangEOGbasedWheelchairRobotic2019} due to their inherently slow behaviour. 
Revising our algorithm to evaluate sight in 3D space instead of 2D could also prevent false positives, disregarding distant gaze points as accidental, but cannot be used for precise filtering due to the high error associated with gaze depth estimation \cite{maimon-morFree3DEndpoint2017}.

\subsubsection{Head-tracking}

While a notable advantage of our approach is the lack of need to perform head-tracking, its addition could remove the restriction of keeping the fiducial markers in the buttons all the time. 
A fiducial-based SLAM approach \cite{Fiducials_ROS} could be used to find the user's head relative to the robot without the need for external systems, allowing it to detect interaction with buttons outside the camera field of view.
Other types of visual markers could be used to improve performance or user experience.  
STag markers could be used to make the system less sensitive to occlusion by other items, as well as provide faster performance and more precise measurements \cite{kalaitzakisFiducialMarkersPose2021}.
The approach deteriorates with distances greater than one meter, but this is not expected to be found in the end application.
Customisable markers such as \textit{Topotags}\cite{yu2021topotag}, could also be used to disguise the markers into the buttons.
Alternatively, markers could be discarded altogether and replaced by image recognition on the interface elements.


\subsubsection{Calibration Drift}


The eye-gaze calibration is susceptible to drift over time, caused due to changing environmental conditions and movement of the glasses relative to the user.
Fujii et al. \cite{fujiiGazeGestureBased2018} demonstrate the use of eye gestures for online gaze calibration to negate drift artefacts.
Calibration is triggered to a predefined sequence of saccadic movements, allowing the user to continuously and invisibly recalibrate during operation by performing frequent tasks. 
Thus, calibration could be triggered by common interactions in our system, such as prolonged looks at the grasping control or alternating between left and right markers.  


\subsubsection{Population Requirements}

While considerations have been suggested for users unable to move their heads, it is not clear if the system would work on individuals with monocular vision. 
The method presented for interaction is limited to people with working binocular vision, as most commercial gaze-tracking systems rely on gaze tracking via eye vergence.
Thus, it may not be suited for patients with monocular vision, reduced ocular control or ocular spasticity, which may be present in patients with cerebral palsy  \cite{TrendsObservedBilateralCerebralPalsy30year_Pizzighello} or stroke. 
While it is not believed that the presented system is suited for all users with serious disabilities, we believe the system offers non-expert users the ability to control, customize and design interfaces to their liking, while keeping the cognitive load low.

\section{Conclusion \& Future Works}
\label{s:Conclusion}


In this document we demonstrate a novel approach for gaze control, discarding screens in favour of physical elements which can be used as a physical interface.  
We apply this approach for the control of a robotic arm in 3D space, demonstrating precision 
and robustness. 
User tests were conducted, where 21 participants demonstrated control of the robot to complete a precision task with minimal training.
A score of $13.71$ out of $16$ was achieved on the Yale-CMU-Berkeley (YCB) Block Pick and Place Protocol, demonstrating dexterity in a precision task over an average execution time of $12$ mins $19$ seconds.
A high score for the System Usability Scale was obtained, placing the system between "Good" and "Excellent" in the SUS adjective rating scale \cite{bangorDeterminingWhatIndividual2009}. 
Similarly, a NASA Task Load index score comparable to more traditional control schemes for assistive robotic devices was obtained. 
None of the demographic factors controlled for were found to affect results, indicating little evidence of digital discrimination. 
Unlike other approaches, we require minimal computational power, no additional hardware for head-tracking, and are not limited by a known selection of items, while using natural-looking hardware. 
Users reported minimal nausea but high intuitiveness, believed to be a result of the screenless GUI approach.


For future work, the main priority is to involve users with disabilities, exploiting the customisation capabilities to design effective strategies. 
The flexible interface opens the door for easy co-design, targeting specific needs, and \textit{Activities of Daily Living}. 
Increasing robustness by supporting external triggers, designing interfaces for intuitive control of rotation, and developing feedback options for the robot arm and interface are of high interest.

%

\appendices

\section{Open Source Packages}
\label{s:OpenSourcePackages}


The final processing pipeline is available under a public GNU license at \cite{Access} as a collection of scripts and a ready-to-use Docker image. 
Instructions and simple scripts to launch the environment, Rviz and rqt\_graph are also provided. 
The system can be quickly integrated with any ROS2 in Docker using the standard \verb|sensor_msgs/msg/Joy| message type.  
The template used for testing can also be found here.
We encourage other researchers to experiment with the gaze-controlled approach.  

The "pure" Docker environments used in this document are also provided, based on Moveit2 and Moveit2 tutorials and for Franka for ROS2, providing a parting point for any general controller.
The Moveit and Franka ROS environments use a large number of nodes to accomplish their tasks and only small pieces of the default code have been changed.
Thus, the images are provided without modifications, with the modified scripts loaded onto the environment and compiled before run-time. 
Since Docker compiling is only needed if changes occur, this allows to alter the behaviour of the large packages with small overhead.

A standalone video summary of the system can be found at \hyperlink{https://youtu.be/hrXuNYLDFds}{https://youtu.be/hrXuNYLDFds} .

\section*{Acknowledgment}

We would like to express our gratitude to the staff at the University of the West of England, the University of Bristol, and Bristol Robotics Laboratory for their technical expertise and support. 
We would like to express our sincere gratitude to all the volunteer participants who have contributed to the result collection and the publication of this research paper.
The first author would like to express his deepest gratitude to his improbably large number of supervisors and their invaluable guidance and continuous support. 
We would like to express our best wishes to little Santiago after being welcomed into this world.

\ifCLASSOPTIONcaptionsoff
  \newpage
\fi



%

\bibliographystyle{ieeetr}
\bibliography{refs}

\end{document}